\documentclass{article}
\usepackage{geometry}
\usepackage[ddmmyyyy]{datetime}
\usepackage[T1]{fontenc}
\usepackage[utf8]{inputenc}

\usepackage{amsthm}
\usepackage{kpfonts}
\usepackage{amsmath}
\usepackage{amssymb}
\usepackage{dsfont}
\usepackage{pifont}

\usepackage{graphicx}
\usepackage[table]{xcolor}
\usepackage{import}

\definecolor{myred}{RGB}{150,0,0}
\definecolor{mygreen}{RGB}{0,150,0}
\definecolor{myblue}{RGB}{0, 101, 189}
\definecolor{myyellow}{RGB}{220, 206, 0}
\definecolor{myorange}{RGB}{255, 153, 51}
\definecolor{mycyan}{RGB}{51, 204, 204}
\definecolor{mypurple}{RGB}{204, 0, 153}

\usepackage{caption}

\usepackage{enumitem}
\usepackage{algorithm}
\usepackage{algorithmicx}
\usepackage{algpseudocode}
\usepackage{booktabs}

\renewcommand{\arraystretch}{1.5}

\usepackage{nicefrac}
\usepackage{bm}
\usepackage{thm-restate}
\usepackage{optidef}
\usepackage{xspace}

\usepackage{amsthm}
\usepackage{thmtools} 
\usepackage{thm-restate}

\declaretheorem[name=Theorem, numberwithin=section]{theorem}
\declaretheorem[name=Definition, style=definition, numberlike=theorem]{definition}
\declaretheorem{lemma, proposition}[style=plain, numberlike=theorem]
\declaretheorem{notation, note, remark}[
  style=plain,
  numberlike=theorem,
]

\usepackage{tikz}
\usepackage{pgfplots}
\usepackage{pgfplotstable}

\usepgfplotslibrary{groupplots}

\tikzstyle{line_node} = [line width=1pt, rounded corners, color=black, ->]
\tikzstyle{line_cv} = [line width=3pt, color=mygreen, line cap=round]

\pgfplotsset{
  lineprimal/.style   = {smooth, myred, very thick},
  linedual/.style     = {cyan, very thick, dashed},
  lineiver/.style  = {smooth, myred, very thick},
  line/.style  = {no markers, smooth, very thick},
}

\pgfplotstableread[col sep = comma]{./data/convergence_toppush.csv}\ConvergenceTopPush
\pgfplotstableread[col sep = comma]{./data/convergence_toppushk.csv}\ConvergenceTopPushK
\pgfplotstableread[col sep = comma]{./data/convergence_patmatnp.csv}\ConvergencePatMat

\pgfmathdeclarefunction{iverson}{1}{%
  \pgfmathparse{max(0, 1 + #1)}%
}
\pgfmathdeclarefunction{hinge}{2}{%
  \pgfmathparse{max(0, 1 + #2 * #1)}%
}
\pgfmathdeclarefunction{quadratic}{2}{%
  \pgfmathparse{(max(0, 1 + #2 * #1))^2}%
}

\usepackage[style=numeric, sorting=nyt]{biblatex}
\newcommand{\Brac}[2][r]{%
  \ifx r#1 \left(       #2 \right)       \else
  \ifx c#1 \left\{      #2 \right\}      \else
  \ifx s#1 \left[       #2 \right]       \else
  \ifx v#1 \left\vert   #2 \right\vert   \else
  \ifx a#1 \left\langle #2 \right\rangle \else
  \ifx t#1 \left\lceil  #2 \right\rceil  \else
  \ifx b#1 \left\lfloor #2 \right\rfloor \else
  \ifx n#1 \left\|      #2 \right\|      \else
  \mathrm{Illegal~option}%
  \fi\fi\fi\fi\fi\fi\fi\fi
}

\newcommand{\clip}[4][s]{
  \ifx s#1 \mathrm{clip}_{\Brac[s]{#2,\; #3}}\Brac{#4} \else
  \ifx u#1 \mathrm{clip}_{\left[#2,\; #3\right)}\Brac{#4} \else
  \ifx l#1 \mathrm{clip}_{\left(#2,\; #3\right]}\Brac{#4} \else
  \mathrm{Illegal~option}%
  \fi\fi\fi
}

\newcommand{\yesmark}{\textcolor{mygreen}{\ding{51}}}%
\newcommand{\nomark}{\textcolor{myred}{\ding{55}}}

\newcommand{\best}[1]{\cellcolor{mygreen!50}#1}
\newcommand{\worst}[1]{\cellcolor{myred!50}#1}

\newcommand{\R}{\mathbb{R}}
\newcommand{\N}{\mathbb{N}}
\newcommand{\X}{\mathbb{X}}

\newcommand{\I}{\mathcal{I}}
\newcommand{\Itil}{\tilde{\mathcal{I}}}
\newcommand{\Ineg}{\I_{-}}
\newcommand{\Ipos}{\I_{+}}

\newcommand{\nall}{n}
\newcommand{\nneg}{n_{-}}
\newcommand{\npos}{n_{+}}
\newcommand{\ntil}{\tilde{n}}

\newcommand{\K}{\mathbb{K}}
\newcommand{\Kall}{\K^{\pm}}
\newcommand{\Kneg}{\K^{-}}

\newcommand{\alphak}{\alpha_{\hat{k}}}
\newcommand{\alphal}{\alpha_{\hat{l}}}
\newcommand{\betak}{\beta_{\hat{k}}}
\newcommand{\betal}{\beta_{\hat{l}}}

\newcommand{\norm}[1]{\Brac[n]{#1}}
\newcommand{\abs}[1]{|#1|}

\newcommand{\Iverson}[1]{\mathds{1}_{\Brac[s]{#1}}}

\newcommand{\Matrix}[1]{\begin{pmatrix} #1 \end{pmatrix}}
\newcommand{\Set}[2]{\Brac[c]{#1 \; \middle\vert \; #2}}
\newcommand{\domain}{\operatorname*{dom}}

\newcommand{\repeatloop}{\texttt{repeat}\xspace}
\newcommand{\forloop}{\texttt{for}\xspace}

\newcommand{\vecab}{\Matrix{\bm{\alpha} \\ \bm{\beta}}}

\newcommand{\TopPush}{\emph{TopPush}\xspace}
\newcommand{\TopPushK}{\emph{TopPushK}\xspace}
\newcommand{\tauFPL}{{\emph{$\tau$-FPL}}\xspace}
\newcommand{\TopMeanK}{\emph{TopMeanK}\xspace}
\newcommand{\PatMat}{\emph{Pat}\&\emph{Mat}\xspace}
\newcommand{\PatMatNP}{{\emph{Pat}\&\emph{Mat-NP}}\xspace}
\newcommand{\Grill}{\emph{Grill}\xspace}
\newcommand{\GrillNP}{\emph{Grill-NP}\xspace}

\newcommand{\SVM}{\emph{SVM}\xspace}

\newcommand{\tpratk}{\operatorname{TPR@}K}
\newcommand{\tpratfpr}{\operatorname{TPR@}\tau}
\newcommand{\auroc}{\operatorname{AUC}}

\usepackage{authblk}

\title{Nonlinear classifiers for ranking problems based on kernelized SVM}
\author[1]{Václav Mácha}
\author[2]{Lukáš Adam}
\author[3]{Václav Šmídl}
\affil[1]{
  Faculty of Nuclear Sciences and Physical Engineering,\protect\\
  Czech Technical University in Prague,\protect\\
  Prague, Czech Republic
  }
\affil[2]{
  Faculty of Electrical Engineering,
  Czech Technical University in Prague,\protect\\
  Prague, Czech Republic
}
\affil[3]{
  Institute of Information Theory and Automation,
  Czech Academy of Sciences,\protect\\
  Prague, Czech Republic
}

\date{}

\begin{document}

\maketitle

\begin{abstract}
    Many classification problems focus on maximizing the performance only on the samples with the highest relevance instead of all samples. As an example, we can mention ranking problems, accuracy at the top or search engines where only the top few queries matter. In our previous work, we derived a general framework including several classes of these linear classification problems. In this paper, we extend the framework to nonlinear classifiers. Utilizing a similarity to SVM, we dualize the problems, add kernels and propose a componentwise dual ascent method.
\end{abstract}

\section{Introduction}

The aim of classical linear binary classification is to separate positive and negative samples by a linear hyperplane. In many applications, it is desirable to separate only a certain number of samples. In such a case, the goal is not to maximize the performance on all samples but only the performance on the required samples with the highest relevance. Such classifiers have many applications. For example, in information retrieval systems, only the most relevant documents should be returned for a given query. Furthermore, they are useful in domains, where a large number of samples needs to be quickly screened and only a small subset of samples needs to be selected for further evaluation.

These problems can be generally written as pushing the positive samples above some decision threshold. The methods differ in the definition of the decision threshold. In our previous work~\cite{adam2021general}, we introduced a general framework that unifies these methods. We showed that several problem classes, which were considered as separate problems so far, fit into the framework. As the most relevant we mention the following methods:
\begin{itemize}
    \item \emph{Ranking problems} focuses on ranking the positive samples higher than the negative ones. Many methods, such as \emph{RankBoost}~\cite{freund2003efficient}, \emph{Infinite Push}~\cite{agarwal2011infinite} or \emph{$p$-norm push}~\cite{rudin2009pnorm} employ a pairwise comparison of samples, which makes them infeasible for larger datasets. This was alleviated in \TopPush~\cite{li2014top} where the authors considered the limit~$p \rightarrow \infty$. Since the $l_{\infty}$ norm from \TopPush is equal to the maximum, the decision threshold from our framework equals to the maximum of scores of negative samples. This was generalized into \TopPushK~\cite{adam2021general} by considering the threshold to be the mean of~$K$ largest scores of negative samples.

     \item Accuracy at the Top~\cite{boyd2012accuracy} focuses on maximizing the number of positive samples above the top $\tau$-quantile of scores. There are many methods on how to solve accuracy at the top. In~\cite{boyd2012accuracy}, the authors assume that the top quantile is one of the samples, construct~$n$ unconstrained optimization problems with fixed thresholds, solve them and select the best solution. This method is computationally expensive. In~\cite{grill2016learning} the authors propose a fast projected gradient descent method. In our previous paper, we proposed a convex approximation of the accuracy at the top called \PatMatNP. This method is reasonably fast and guaranteed the existence of global optimum.
\end{itemize}

The deficiency of methods from this framework is that they usually cover only linear classifiers. However, as many problems are not linearly separable, nonlinear classifiers are needed. In this work, we show how to extend our framework into nonlinear classification problems. To do so, we use the fact that our framework is similar to the primal formulation of support vector machines~\cite{cortes1995support}. The classical way to incorporate nonlinearity into SVM is to derive the dual formulation~\cite{boyd2004convex} and to employ the kernels method~\cite{scholkopf2001learning}. In this work, we follow this approach, derive dual formulations for the considered problems and add nonlinear kernels to them. Moreover, as dual problems are generally expensive to solve, we derive a quick method to solve them. This is a modification of the coordinate-wise dual ascent from~\cite{hsieh2008dual}. For a review of other approaches see \cite{batmaz2019review,werner2019review}.

The paper is organized as follows: In Section~\ref{sec: framework} we recall the unified framework derived in~\cite{adam2021general}. In Section  and two class of problems that falls into it. Moreover, for selected methods, we derive their dual formulations. Namely, we focus on \TopPush, \TopPushK and \PatMatNP. In Section~\ref{sec: kernels}, we show how to add nonlinear kernels into dual formulations. In Section~\ref{sec: coordinate descent} derive a new method for solving these dual problems and perform its complexity analysis. Since our method depends on the chosen problem and surrogate function, we provide a concrete form of the solution for \TopPushK with the quadratic hinge loss. Solutions for other problems are provided in Appendix. Finally, in Section~\ref{sec: experiments} we present the description of performance criteria, choice of hyperparameters and description of datasets. The rest of the section is focused on the results of numerical experiments.

% ------------------------------------------------------------------------------
%  Framework
% ------------------------------------------------------------------------------
\section{Framework}\label{sec: framework}

In this section, we recall the general framework for classification at the top introduced in~\cite{adam2021general}. For simplicity, we use the following notation in the rest of the text.

\begin{notation}[Dataset]\label{not: dataset}
  In this work, we use label~$0$ to encode the negative class and label~$1$ to encode the positive class. By a dataset of size~$n \in \N$ we mean a set of pairs in the following form
  \begin{equation*}
    \mathcal{D} = \Brac[c]{(\bm{x}_i, y_i)}_{i=1}^{n},
  \end{equation*}
  where~$\bm{x}_i \in \R^d$ represents samples and~$y_i \in \{0, 1\}$ corresponding labels. To simplify future notation, we denote a set of all indices of dataset~$\mathcal{D}$ as~$\I = \Ineg \cup \Ipos,$ where
  \begin{equation*}
    \begin{aligned}
      \Ineg & = \Set{i}{i \in \{1, 2, \ldots, n\} \; \land \; y_i = 0}, \\
      \Ipos & = \Set{i}{i \in \{1, 2, \ldots, n\} \; \land \; y_i = 1}.
    \end{aligned}
  \end{equation*}
  We also denote the number of negative samples in~$\mathcal{D}$ as~$\nneg = \Brac[v]{\Ineg}$ and the number of positive samples in~$\mathcal{D}$ as~$\npos = \Brac[v]{\Ipos}.$ The total number of samples is~$n = \nneg + \npos.$ 
\end{notation}

Linear binary classification is a problem of finding a linear hyperplane that separates a group of positive samples from a group of negative samples and achieves the lowest possible error. For a sample~$\bm{x} \in \R^d,$ the prediction for a linear classifier amounts to
\begin{equation*}
    \bm{x} \textnormal{ has }
    \begin{dcases*}
        \textnormal{positive label} & if~$\bm{w}^{\top} \bm{x} \geq t$, \\
        \textnormal{negative label} & otherwise.
    \end{dcases*}
\end{equation*}
Here,~$\bm{w} \in \R^{d}$ is the normal vector to the separating hyperplane and~$t \in \R$ is a decision threshold. The well-known example of such a classifier is a support vector machine~\cite{cortes1995support} where the decision threshold~$t$ is a free variable. However, many important binary classification problems maximize the performance only for a certain amount of samples with the highest scores~$s = \bm{w}^{\top}\bm{x}.$ In these cases, the threshold~$t$ is not a free variable but a function of the scores. In our previous work~\cite{adam2021general}, we formulated a general framework for maximizing performance above the threshold $t$ as
\begin{mini}{\bm{w}}{
  C_1 \cdot \sum_{i \in \Ineg}\Iverson{s_i \geq t} + C_2 \cdot \sum_{i \in \Ipos}\Iverson{s_i < t}
}{\label{eq: aatp counts}}{}
  \addConstraint{s_i}{= \bm{w}^{\top}\bm{x}_i, \quad}{i \in \I}
  \addConstraint{t}{= G\Brac{\bm{s}, \bm{y}},}
\end{mini}
where function~$G \colon \R^n \times \{0, 1\}^n \to \R$ takes the scores and labels of all samples and computes the decision threshold and~$\Iverson{\cdot{}}$ is the Iverson function which is used to count misclassified samples and is defined as
\begin{equation}\label{eq: iverson}
  \Iverson{x} = \begin{cases}
    0 & \quad \text{if } x \text{ is false}, \\
    1 & \quad \text{if } x \text{ is true}.
  \end{cases}
\end{equation}
The concrete form of the function~$G$ that defines the decision threshold depends on the used problem. Note the important distinction from the standard binary classification: the decision threshold is no longer fixed (as in the case of neural networks) or trained independently (as in SVM) but is a function of scores of all samples. Therefore, the minimization in problem~\eqref{eq: aatp counts} is performed only concerning the one variable~$\bm{w}.$

The objective function in~\eqref{eq: aatp counts} is a weighted sum of false-positive and false-negative counts. Since these counts are discontinuous due to the presence of the Iverson function, the whole objective function is discontinuous too. Therefore, problem~\eqref{eq: aatp counts} is difficult to solve. One way how to simplify the problem is to derive its continuous approximation. The usual approach is to employ a surrogate function to replace the Iverson function~\cite{li2014top, grill2016learning}.

\begin{notation}[Surrogate function]\label{not: surrogates}
  To approximate the Iverson function~\eqref{eq: iverson}, we use any surrogate function~$l$ that is convex, non-negative, and non-decreasing with~$l(0) = 1,$ and~$l(s) \to 0$ as~$s \to -\infty.$ As examples of such function, we can mention the hinge loss or the quadratic hinge loss defined by
  \begin{align*}
    l_{\text{hinge}}(s) & = \max\Brac[c]{0, 1 + s}, &
    l_{\text{quadratic}}(s) & = \Brac{\max\Brac[c]{0, 1 + s}}^2.
  \end{align*}
  Figure~\ref{fig: surrogates} compares the Iverson function with the hinge and quadratic hinge loss with scaled inputs by~$\vartheta = 2$ and without scaling. We use~$\vartheta > 0$ to denote any scaling parameter.
\end{notation}

\begin{figure}[t]
  \centering
  \begin{tikzpicture}
    \begin{axis}[
      height=6cm,
      width=14cm,
      xmin = -2,
      xmax = 2,
      ymin = -0.1,
      ymax = 4,
      domain = -2:2,
      smooth,
      xtick = {-2, -1, 0, 1, 2},
      ytick = {0, 1, 2, 3, 4},
      xlabel={$s$},
      ylabel={$l(s)$},
      grid=both,
      major grid style={
        dotted,
        gray,
      },
      legend pos =north west,
      legend cell align={left},
      enlargelimits = false,
    ]
    
      \addplot[lineiver, no markers, forget plot] coordinates {(-2, 0) (0, 0)};
      \addplot[lineiver, mark=*, mark options={fill=white}, forget plot] coordinates {(0, 0)};
      \addplot[lineiver, mark=*, mark options={fill=myred}, forget plot] coordinates {(0, 1)};
      \addplot[lineiver, no markers] coordinates {(0, 1) (2.1, 1)};
      \addlegendentry{$\Iverson{s \geq 0}$}
      \addplot [line, cyan] {hinge(x, 1)};
      \addlegendentry{$l_{\text{hinge}}(s)$}
      \addplot [line, myorange, dashed] {hinge(x, 2)};
      \addlegendentry{$l_{\text{hinge}}(2s)$}
      \addplot [line, myblue, dotted] {quadratic(x, 1)};
      \addlegendentry{$l_{\text{quadratic}}(s)$}
      \addplot [line, mygreen, dashdotted] {quadratic(x, 2)};
      \addlegendentry{$l_{\text{quadratic}}(2s)$}
    \end{axis}
  \end{tikzpicture}
  \caption{Comparison of the approximation quality of the Iverson function using different surrogate functions and scaling parameters.}
  \label{fig: surrogates}
\end{figure}

By replacing the Iverson function in the objective function of~\eqref{eq: aatp counts} with its surrogate approximation and adding a regularization for better numerical stability, we get
\begin{mini}{\bm{w}}{
  \frac{\lambda}{2} \norm{\bm{w}}^2 + C_1 \cdot \sum_{i \in \Ineg}l(s_i - t) + C_2 \cdot \sum_{i \in \Ipos}l(t - s_i)
  }{\label{eq: aatp surrogate}}{}
  \addConstraint{s_i}{= \bm{w}^{\top}\bm{x}_i, \quad i \in \I}
  \addConstraint{t}{= G\Brac{\bm{s}, \bm{y}}.}
\end{mini}
The resulting objective function is continuous, and therefore the problem is easier to solve than the original problem~\eqref{eq: aatp counts}.

As we derived in~\cite{adam2021general}, there are many problems belonging to the general framework~\eqref{eq: aatp counts}. The summary of all formulations is provided in Table~\ref{tab: summary formulations}. However, this framework handles only linear classification problems. As many problems are not linearly separable, this is often not sufficient. To generalize the framework to nonlinear classifiers, we realize that~\eqref{eq: aatp surrogate} is similar to the primal formulation of the SVM~\cite{cortes1995support}. We will follow the standard way to incorporate nonlinearity into SVM by deriving the dual problem~\cite{boyd2004convex} and using the kernels methods~\cite{scholkopf2001learning}.

In the next section, we introduce two problem families based on formulations from~\cite{adam2021general} and for each of them, we derive its dual formulation. Namely, we will discuss family of \TopPushK formulations and family of \PatMat formulations.

\begin{table}
  \centering
  \begin{tabular}{lcccccc}
    \hline
    \textbf{Formulation}
      & \textbf{Source}
      & \textbf{Ours}
      & \textbf{Hyper-parameters}
      & $C_1$
      & $C_2$
      & \textbf{Threshold} \\
    \hline
    \TopPush
      & \cite{li2014top}
      & \nomark
      & $\lambda$
      & 0
      & $\frac{1}{\npos}$
      & $s_{[1]}^-$ \\
    \TopPushK
      & \cite{adam2021general}
      & \yesmark
      & $\lambda,$ $K$
      & 0
      & $\frac{1}{\npos}$
      & $\frac{1}{K}\sum_{i = 1}^{K} s_{[i]}^-$ \\
    \hline
    \Grill
      & \cite{grill2016learning}
      & \nomark
      & $\lambda$
      & $\frac{1}{\nneg}$
      & $\frac{1}{\npos}$
      & $\max \Set{t}{\frac{1}{\nall} \sum_{i \in \I} \Iverson{s_i \geq t} \geq \tau}$ \\
    \TopMeanK
      & ---
      & \nomark
      & $\lambda$
      & 0
      & $\frac{1}{\npos}$
      & $\frac{1}{K} \sum_{i=1}^{K} s_{[i]}$ \\
    \PatMat
      & \cite{adam2021general}
      & \yesmark
      & $\lambda,$ $\vartheta$
      & 0
      & $\frac{1}{\npos}$
      & $\frac{1}{\nall} \sum_{i \in \I} l\Brac{\vartheta(s_i - t)} = \tau$ \\
    \hline
    \GrillNP
      & ---
      & \nomark
      & $\lambda$
      & $\frac{1}{\nneg}$ 
      & $\frac{1}{\npos}$
      & $\max \Set{t}{ \frac{1}{\nneg} \sum_{i \in \Ineg} \Iverson{s_i \geq t} \geq \tau}$ \\
    \tauFPL
      & \cite{zhang2018tau}
      & \nomark
      & $\lambda$
      & 0
      & $\frac{1}{\npos}$
      & $\frac{1}{\nneg\tau} \sum_{i=1}^{\nneg\tau} s^-_{[i]}$ \\
    \PatMatNP
      & \cite{adam2021general}
      & \yesmark
      & $\lambda,$ $\vartheta$
      & 0
      & $\frac{1}{\npos}$
      & $\frac{1}{\nneg} \sum_{i \in \Ineg} l\Brac{\vartheta(s_i - t)} = \tau$ \\
    \hline
  \end{tabular}
  \caption{Summary of problem fomrulations that fall in the framework~\eqref{eq: aatp surrogate}. Column \textbf{Formulation} shows the name of the formulation that we use in this work. Column \textbf{Source} is the citation of the work where the formulation was introduced. Column \textbf{Ours} shows whether the formulation was introduced in any of our previous papers. Column \textbf{Hyperparameters} shows the hyperparameters available for each formulation. The last three columns show the values of parameters~$C_1,$~$C_2$ and the form of the decision threshold for given framework~\eqref{eq: aatp surrogate}.}
  \label{tab: summary formulations}
\end{table}

% ------------------------------------------------------------------------------
% Formulation of dual problems
% ------------------------------------------------------------------------------
\section{Derivation of Dual Problems}\label{sec:Derivation of dual problems}

In Section~\ref{sec: framework}, we introduced a general framework for binary classification at the top. Moreover, we showed that several problem classes, considered separate problems so far, fit into this framework. Many formulations have nice theoretical properties such as convexity or differentiability in this specific case. However, many real-world problems are not linearly separable, and in such cases, the approach from the previous section is not sufficient. In this section, we use the similarity of~\eqref{eq: aatp surrogate} to primal formulation of SVM~\cite{cortes1995support} and derive dual forms for almost all formulations from Table~\ref{tab: summary formulations}. Then we use the kernel method~\cite{scholkopf2001learning} to introduce nonlinearity into the dual formulations. Moreover, as dual problems are generally computationally expensive, we propose an efficient method to solve them.

This section is dedicated to deriving dual forms for almost all formulations from Table~\ref{tab: summary formulations}. We do not discuss \Grill and \GrillNP formulations in the following text since both formulations are not convex, and therefore their primal and dual formulations are not equivalent. Since many of the remaining formulations are very similar, we divide them into two families:
\begin{itemize}
  \item \textbf{\TopPushK family:} \TopPush, \TopPushK, \TopMeanK and \tauFPL.
  \item \textbf{\PatMat family:} \PatMat and \PatMatNP.
\end{itemize}
Both families use surrogate false-negative rate as an objective function. Moreover, all formulations from \TopPushK family use the mean of~$K$ highest scores of all or negative samples as a threshold and differ only in the definition of~$K.$ Finally, both formulations from \PatMat family use a surrogate approximation of the top $\tau$-quantile of scores of all or negative samples. In other words, we have two families of formulations that share the same objective function and the same form of the decision threshold. Therefore, we derive all results for the general form of these two families. Before we start, we need to introduce the concept of conjugate functions.

\begin{definition}[Conjugate function~\cite{boyd2004convex}]\label{def: conjugate}
  Let~$l \colon \R^n \to \R.$ The function~$l^{\star} \colon \R^n \to \R,$ defined as
  \begin{equation*}
    l^{\star} (\bm{y})
      =  \sup_{\bm{x} \in \domain l} \{\bm{y}^{\top}\bm{x} - l(\bm{x})\}.
  \end{equation*}
  is called the conjugate function of~$l.$ The domain of the conjugate function
  consists of $\bm{y} \in \R^n$ for which the supremum is finite. 
\end{definition}

These functions will play a crucial role in the resulting form of dual problems. Recall the hinge loss and quadratic hinge loss function defined in Notation~\ref{not: surrogates}
\begin{align*}
    l_{\text{hinge}}(s) & = \max\Brac[c]{0, 1 + s}, &
    l_{\text{quadratic}}(s) & = \Brac{\max\Brac[c]{0, 1 + s}}^2.
\end{align*}
The conjugate function for the hinge loss can be found in~\cite{shnlev2014accelerated} and has the following form
\begin{equation}\label{eq: conjugate hinge}
  l_{\text{hinge}}^{\star}(y) =
  \begin{cases}
    -y & \text{if } y \in [0, 1], \\
    \infty & \text{otherwise.}
  \end{cases}  
\end{equation}
Similarly, the conjugate function for the quadratic hinge was computed in~\cite{kanamori2013conjugate} as
\begin{equation}\label{eq: conjugate quadratic hinge}
  l_{\text{quadratic}}^{\star}(y) =
  \begin{cases}
    \frac{y^2}{4} - y & \text{if } y \geq 0, \\
    \infty & \text{otherwise.}
  \end{cases}
\end{equation}

\begin{notation}[Kernel Matrix]\label{not: kernel matrix}
  To simplify the future notation, we introduce matrix~$\X$ of all samples. Each row of~$\X$ represents one sample and is defined for all~$i \in \I$ as
  \begin{equation*}
    \X_{i, \bullet} = \bm{x}_i^{\top}.
  \end{equation*}
  In the same way, we defined matrices~$\X^+,$~$\X^-$ of all negative and positive samples with rows defined as
  \begin{align*}
    \X^{-}_{i, \bullet} & = \bm{x}_i^{\top} \quad i = 1, \;, 2, \ldots, \; n^-, \\
    \X^{+}_{i, \bullet} & = \bm{x}_i^{\top} \quad i = 1, \;, 2, \ldots, \; n^+.
  \end{align*}
  Moreover, for all formulations that use only negative samples to compute the threshold~$t$, we define kernel matrix~$\Kneg$ as
  \begin{equation*}
    \Kneg = \Matrix{\X^+ \\ - \X^-} \Matrix{\X^+ \\ - \X^-}^\top = \Matrix{\X^+ \X^{+\top} & -\X^+ \X^{-\top} \\ -\X^- \X^{+\top} & \X^- \X^{-\top} }.
  \end{equation*}
  and for all formulations that use only all samples to compute the threshold~$t$, we define kernel matrix~$\Kall$ as
  \begin{equation*}
    \Kall = \Matrix{\X^+ \\ - \X} \Matrix{\X^+ \\ - \X}^\top = \Matrix{\X^+ \X^{+\top} & -\X^+ \X^{\top} \\ -\X \X^{+\top} & \X \X^{\top} }.
  \end{equation*}
  In the rest of the text, matrix~$\K$ always refers to one of the kernel matrices defined above. 
\end{notation}

\subsection{Family of \TopPushK Formulations}

In this section, we focus on the family of \TopPushK formulations. The general optimization problem that covers all formulations from this family can be written in the following way
\begin{mini!}{\bm{w}}{
  \frac{1}{2} \norm{\bm{w}}^2 + C \sum_{i \in \Ipos} l(t - \bm{w}^{\top} \bm{x}_i)
  }{\label{eq: toppushk family}}{\label{eq: toppushk family L}}
  \addConstraint{s_j}{= \bm{w}^{\top} \bm{x}_j, \quad j \in \Itil \label{eq: toppushk family c1}}
  \addConstraint{t}{= \frac{1}{K} \sum_{j = 1}^{K} s_{[j]}, \label{eq: toppushk family c2}}
\end{mini!}
where~$C \in \R.$ The set of indices~$\Itil$ equals~$\I$ for \TopMeanK and~$\Ineg$ for other formulations. The parameter~$K$ equals~$1$ for \TopPush, $K$ for \TopPushK, $\nall \tau$ for \TopMeanK, and $\nneg \tau$ for \tauFPL. Note that we use an alternative formulation with constant~$C$, since it is more similar to the standard SVM, and we wanted to stress this similarity. For~$C = \nicefrac{1}{\lambda \npos}$ the new formulation is identical to the original one.

The following theorem shows the dual form of formulation~\eqref{eq: toppushk family}. The dual formulation for \TopPush was originally derived in~\cite{li2014top}. We only show, that our general dual formulation also covers this special case. To keep the readability as simple as possible, we postpone all proofs to Appendix.

\begin{restatable}[Dual formulation for \TopPushK family]{theorem}{topdual}\label{thm: toppushk family dual}
  Consider Notation~\ref{not: kernel matrix}, surrogate function~$l,$ and formulation~\eqref{eq: toppushk family}. Then the corresponding dual problem has the following form
  \begin{maxi!}{\bm{\alpha}, \bm{\beta}}{
    - \frac{1}{2} \vecab^\top \K \vecab
    - C \sum_{i = 1}^{\npos} l^{\star}\Brac{\frac{\alpha_i}{C}}
    }{\label{eq: toppushk family dual}}{\label{eq: toppushk family dual L}}
    \addConstraint{\sum_{i = 1}^{\npos} \alpha_i}{= \sum_{j = 1}^{\ntil} \beta_j \label{eq: toppushk family dual c1}}
    \addConstraint{0 \leq \beta_j}{\leq \frac{1}{K} \sum_{i = 1}^{\npos} \alpha_i, \quad j = 1, 2, \ldots, \ntil, \label{eq: toppushk family dual c2}}
  \end{maxi!}
  where~$l^{\star}$ is conjugate function of~$l$ and
  \begin{center}
    \renewcommand*{\arraystretch}{1}
    \begin{tabular}{lcccc}
        & $K$
        & $\K$
        & $\ntil$
        & $\tilde{\bm{x}}_j$ \\
      \hline
      \TopPush
        & $1$
        & $\Kneg$
        & $\nneg$
        & $\bm{x}^-_j$ \\
      \TopPushK
        & $K$
        & $\Kneg$
        & $\nneg$
        & $\bm{x}^-_j$ \\
      \TopMeanK
        & $\nall \tau$
        & $\Kall$
        & $\nall$
        & $\bm{x}_j$ \\
      \tauFPL
        & $\nneg \tau$
        & $\Kneg$
        & $\nneg$
        & $\bm{x}^-_j$ \\
    \end{tabular}
  \end{center}
  If~$K = 1,$ the upper bound in the second constraint~\eqref{eq: toppushk family dual c2} vanishes due to the first constraint. Finally, the primal variables~$\bm{w}$ can be computed from dual variables as follows
  \begin{equation}\label{eq: toppushk family dual to primal}
    \bm{w} = \sum_{i = 1}^{\npos} \alpha_i \bm{x}^+_i - \sum_{j = 1}^{\ntil} \beta_j \tilde{\bm{x}}_j.
  \end{equation}
\end{restatable}

\subsection{Family of \PatMat Formulations}

In the same way, as for \TopPushK family, we introduce a general optimization problem that covers all formulations from \PatMat family and reads
\begin{mini}{\bm{w}}{
  \frac{1}{2} \norm{\bm{w}}^2 + C \sum_{i \in \Ipos} l(t - \bm{w}^{\top} \bm{x}_i)
  }{\label{eq: patmat family}}{}
  \addConstraint{t}{\;\; \text{solves} \;\; \frac{1}{\ntil}\sum_{i \in \Itil} l\Brac{\vartheta(\bm{w}^{\top} \bm{x}_j - t)} = \tau,}
\end{mini}
where~$C \in \R.$ For \PatMat we have~$\Itil = \I$ and~$\ntil = \nall.$ For \PatMatNP we have~$\Itil = \Ineg$ and~$\ntil = \nneg.$ Again, we use the alternative formulation with constant~$C.$ The following theorem shows the dual form of the formulation~\eqref{eq: patmat family}.

\begin{restatable}[Dual formulation for \PatMat family]{theorem}{patdual}\label{thm: patmat family dual}
  Consider Notation~\ref{not: kernel matrix}, surrogate function~$l,$ and formulation~\eqref{eq: patmat family}. Then the corresponding dual problem has the following form
  \begin{maxi!}{\bm{\alpha}, \bm{\beta}, \delta}{
    - \frac{1}{2} \vecab^\top \K \vecab
    - C \sum_{i = 1}^{\npos} l^{\star}\Brac{\frac{\alpha_i}{C}}
    - \delta \sum_{j = 1}^{\ntil} l^{\star} \Brac{\frac{\beta_j}{\delta\vartheta }}
    - \delta \ntil \tau
    }{\label{eq: patmat family dual}}{\label{eq: patmat family dual L}}
    \addConstraint{\sum_{i = 1}^{\npos} \alpha_i}{= \sum_{j = 1}^{\ntil} \beta_j \label{eq: patmat family dual c1}}
    \addConstraint{\delta }{\geq 0, \label{eq: patmat family dual c2}}
  \end{maxi!}
  where~$l^{\star}$ is conjugate function of~$l,$~$\vartheta > 0$ is a scaling parameter and
  \begin{center}
    \renewcommand*{\arraystretch}{1}
    \begin{tabular}{lccc}
        & $\K$
        & $\ntil$
        & $\tilde{\bm{x}}_j$ \\
      \hline
      \PatMat
        & $\Kall$
        & $\nall$
        & $\bm{x}_j$ \\
      \PatMatNP
        & $\Kneg$
        & $\nneg$
        & $\bm{x}^-_j$ \\
    \end{tabular}
  \end{center}
  Finally, the primal variables~$\bm{w}$ can be computed from dual variables as follows
  \begin{equation}\label{eq: patmat family dual to primal}
    \bm{w} = \sum_{i = 1}^{\npos} \alpha_i \bm{x}^+_i - \sum_{j = 1}^{\ntil} \beta_j \tilde{\bm{x}}_j.
  \end{equation}
\end{restatable}

\begin{note}
  For simplicity, the rest of the section covers only the \TopPushK formulation with hinge loss. We use this formulation since it is the prototypical example for the \TopPushK family of formulations. The results for the rest of the formulations from this family can be derived almost identically. Moreover, results for the \PatMat family of formulations can be derived similarly. Therefore, derivations for the \TopPushK family with quadratic hinge loss and the \PatMat family with hinge and quadratic hinge loss are postponed to Appendix.
\end{note}

\subsection{Kernels}\label{sec: kernels}

As we mentioned at the beginning of the section, our goal is to extend our framework to be usable for linearly inseparable problems. In two previous sections, we derived dual formulations for \TopPushK and \PatMat families. In this section, we show how to employ the kernels method~\cite{scholkopf2001learning} to introduce nonlinearity into these dual formulations. For simplicity, we focus only on the \TopPushK formulation that computes the decision threshold only from negative samples. As mentioned in Notation~\ref{not: kernel matrix}, \TopPushK formulation uses kernel matrix~$\K = \Kneg.$ The following derivation is the same for all other formulations.

To add kernels, we first realize that primal variables~$\bm{w}$ can be computed from dual variables~$\bm{\alpha},$~$\bm{\beta}$ using~\eqref{eq: toppushk family dual to primal}. Therefore, the classification score for any sample~$\bm{x}$ can be calculated as follows
\begin{equation}\label{eq:pred_linear}
  s
    = \bm{w}^{\top} \bm{x}
    = \sum_{i = 1}^{\npos} \alpha_i \bm{x}^{\top} \bm{x}_i^+ - \sum_{i = 1}^{\nneg} \beta_i \bm{x}^{\top} \bm{x}_i^-.
\end{equation}
Importantly, all samples~$\bm{x}_i$ in the previous formula occur only in the dot product with~$\bm{x}$ and not separately. This property allows us to use the standard kernel trick from SVMs~\cite{cortes1995support}. The kernel trick replaces the dot product of the vectors from input space using the so-called kernel function~$k: \R^d \times \R^d \to \R.$ This function represents a dot product in the space of a higher dimension
\begin{equation*}
  k(\bm{x}, \bm{x}') = \phi(\bm{x})^{\top} \phi(\bm{x}'),
\end{equation*}
where~$\phi: \R^d \to \R^D$ is a mapping function. The idea is to transform the input vectors using~$\phi$ into some feature space in which the classification problem is easier to solve. However, getting the explicit formula for the mapping function is usually very hard. The kernel trick allows us to avoid this explicit mapping to the feature space since we can only replace the dot product in~\eqref{eq:pred_linear} by the kernel function~$k$
\begin{equation}\label{eq: dual scores kernel}
  s = \sum_{i = 1}^{\npos} \alpha_i k\Brac{\bm{x}, \bm{x}^+_i} - \sum_{i = 1}^{\nneg} \beta_i k\Brac{\bm{x}, \bm{x}^-_i}.
\end{equation}
The downside of this approach is, that we can not compute the primal variables using~\eqref{eq: toppushk family dual to primal} if we do not know the mapping function~$\phi.$ We always have to calculate the scores using the formula above, which is computationally expensive.

Now we must show how to modify the original dual problem~\eqref{eq: toppushk family dual} to incorporate kernels. Recall the form of the kernel matrix~$\K$ for \TopPushK
\begin{equation*}
  \K
    = \Matrix{\X^+ \X^{+\top} & -\X^+ \X^{-\top} \\ -\X^- \X^{+\top} & \X^- \X^{-\top}}.
\end{equation*}
Since each component of the kernel matrix~$\K$ is computed as a dot product of two training samples, we can replace~$\K$ with a matrix in the following form
\begin{equation}\label{eq: kernel matrix nonlinear}
  \K = \Matrix{k\Brac{\X^+, \X^{+}} & -k\Brac{\X^+, \X^{-}} \\ -k\Brac{\X^-, \X^{+}} & k\Brac{\X^-, \X^{-}}}.
\end{equation}
The kernel function~$k(\cdot,\; \cdot)$ is applied to all rows of both arguments. In other words, if we use the kernel trick, the original dual problem~\eqref{eq: toppushk family dual} remains almost the same. The only change is in the construction of the kernel matrix.

\subsection{Coordinate Descent Algorithm}\label{sec: coordinate descent}

In the previous sections, we derived dual formulations for \TopPushK and \PatMat families of formulations. Moreover, we showed how to incorporate non-linear kernels into these formulations. As a result, we can use all presented formulations even for linearly non-separable problems. However, the dimension of the dual problems is at least equal to the number of all samples~$n,$ and therefore, it is computationally expensive to use standard techniques such as gradient descent. To handle this issue, the standard coordinate descent algorithm~\cite{chang2008coordinate, hsieh2008dual} has been proposed in the context of SVMs. In this section, we derive a coordinate descent algorithm suitable for our dual problems~(\ref{eq: toppushk family dual},~\ref{eq: patmat family dual}). We also show that we can reduce the whole optimization problem to a one-dimensional quadratic optimization problem with a closed-form solution in every iteration. Therefore, every iteration of our algorithm is cheap. For a review of other approaches see~\cite{batmaz2019review,werner2019review}. 

Recall that we perform all derivations only for \TopPushK with hinge loss. Classification scores can be computed directly from dual variables as shown in~\eqref{eq: dual scores kernel}. Using the definition~\eqref{eq: kernel matrix nonlinear} of kernel matrix~$\K$, we can define a vector of scores~$\bm{s}$ by
\begin{equation}\label{eq: dual scores}
  \bm{s} = \K \vecab.
\end{equation}
Note that dual scores are not identical to the primal ones~\eqref{eq:pred_linear} (even though we use the same notation). The main difference is that dual scores use kernel function~$k.$ Therefore, they are equivalent only if the kernel function is defined as a dot product in the input space, i.e., if~$k(\bm{x}, \bm{x}') = \bm{x}^{\top} \bm{x}'.$ To simplify the indexing of the vector of scores~\eqref{eq: dual scores} and kernel matrix~$\K$, we introduce a new notation in Notation~\ref{not: dual update rules}.

\begin{notation}\label{not: dual update rules}
  Consider any index~$l$ that satisfies~$1 \leq l \leq \npos + \ntil.$ Note that the length of dual variable~$\bm{\alpha}$ is~$\npos$ for both formulations~\eqref{eq: toppushk family dual} and~\eqref{eq: patmat family dual}. Therefore, we can define auxiliary index~$\hat{l}$ as 
  \begin{equation*}
    \hat{l} = \begin{cases}
      l & \text{if } l \leq \npos, \\
      l - \npos & \text{otherwise}.
    \end{cases}
  \end{equation*}
  Then the index~$l$ can be safely used for kernel matrix~$\K$ or vector of scores~$\bm{s},$ while its corresponding version~$\hat{l}$ can be used for dual variables~$\bm{\alpha}$ or~$\bm{\beta}.$
\end{notation}

\subsubsection{Update Rules}\label{sec: Top coordinate descent}

Consider dual formulation~\eqref{eq: toppushk family dual} from Theorem~\ref{thm: toppushk family dual} and fixed feasible dual variables~$\bm{\alpha},$~$\bm{\beta}.$ Our goal in this section is to derive an efficient iterative procedure for solving this problem. We follow the ideas presented in~\cite{chang2008coordinate, hsieh2008dual} for solving SVMs using a coordinate descent algorithm. However, we must modify the approach since we have an additional constraint~\eqref{eq: toppushk family dual c1}.  Due to this constraint, we always have to update (at least) two components of dual variables~$\bm{\alpha},$~$\bm{\beta}.$ There are only three update rules which modify two components of~$\bm{\alpha},$~$\bm{\beta},$ and satisfy constraints~\eqref{eq: toppushk family dual c1}. The first one updates two components of~$\bm{\alpha}$
\begin{subequations}\label{eq: update rules}
\begin{align}\label{eq: update rule a,a}
  \alphak & \to \alphak + \Delta, & \quad
  \alphal & \to \alphal - \Delta, & \quad
  \bm{s} & \to \bm{s} + \Brac{\K_{\bullet, k} - \K_{\bullet, l}}\Delta,
\end{align}
where~$\K_{\bullet, i}$ denotes $i$-th column of~$\K$ and indices~$\hat{k},$~$\hat{l}$ are defined in Notation~\ref{not: dual update rules}. Note that the update rule for~$\bm{s}$ does not use matrix multiplication but only vector addition. The second rule updates one component of~$\bm{\alpha}$ and one component of~$\bm{\beta}$ 
\begin{align}\label{eq: update rule a,b}
  \alphak & \to \alphak + \Delta, & \quad
  \betal  & \to \betal  + \Delta, & \quad
  \bm{s} & \to \bm{s} + \Brac{\K_{\bullet, k} + \K_{\bullet, l}}\Delta,
\end{align}
and the last one updates two components of~$\bm{\beta}$
\begin{align}\label{eq: update rule b,b}
  \betak & \to \betak + \Delta, & \quad
  \betal & \to \betal - \Delta, & \quad
  \bm{s}  & \to \bm{s} + \Brac{\K_{\bullet, k} - \K_{\bullet, l}}\Delta.
\end{align}
\end{subequations}
Using any of the update rules above, the problem~\eqref{eq: toppushk family dual} can be written as a one-dimensional quadratic problem in the following form
\begin{maxi*}{\Delta}{
  -\frac{1}{2} a(\bm{\alpha}, \bm{\beta}) \Delta^2
  - b(\bm{\alpha}, \bm{\beta}) \Delta
  - c(\bm{\alpha}, \bm{\beta})
  }{}{}
  \addConstraint{\Delta_{lb}(\bm{\alpha}, \bm{\beta})}{\leq \Delta \leq \Delta_{ub}(\bm{\alpha}, \bm{\beta})}
\end{maxi*}
where~$a,$~$b,$~$c,$~$\Delta_{lb},$~$\Delta_{ub}$ are constants with respect to~$\Delta.$ The optimal solution to this problem is
\begin{equation}\label{eq: Delta optimal}
  \Delta^{\star} = \clip{\Delta_{lb}}{\Delta_{ub}}{\gamma},
\end{equation}
where~$\gamma = -\frac{b}{a}$ and~$\clip{a}{b}{x}$ amounts to clipping (projecting)~$x$ to interval~$[a, b].$ Since we assume one of the update rules~\eqref{eq: update rules}, the constraint~\eqref{eq: toppushk family dual c1} is always satisfied after the update. Even though all three update rules hold for any surrogate, the calculation of the optimal~$\Delta^{\star}$ depends on the concrete form of surrogate function. In the following text, we show the closed-form formula for~$\Delta^{\star},$ when the hinge loss function from Notation~\ref{not: surrogates} is used. 

Plugging the conjugate~\eqref{eq: conjugate hinge} of the hinge loss into the dual formulation~\eqref{eq: toppushk family dual} yields
\begin{maxi!}{\bm{\alpha}, \bm{\beta}}{
  - \frac{1}{2} \vecab^\top \K \vecab
  + \sum_{i = 1}^{\npos} \alpha_i
  }{\label{eq: Top dual hinge}}{\label{eq: Top dual hinge L}}
  \addConstraint{\sum_{i = 1}^{\npos} \alpha_i}{= \sum_{j = 1}^{\ntil} \beta_j
  \label{eq: Top dual hinge c1}}
  \addConstraint{0 \leq \alpha_i}{\leq C,}{i = 1, 2, \ldots, \npos
  \label{eq: Top dual hinge c2}}
  \addConstraint{0 \leq \beta_j}{\leq \frac{1}{K} \sum_{i = 1}^{\npos} \alpha_i, \quad}{j = 1, 2, \ldots, \ntil.
  \label{eq: Top dual hinge c3}}
\end{maxi!}
The form of~$\K$ and~$\ntil$ depends on the used formulation as discussed in Theorem~\ref{thm: toppushk family dual}. Moreover, the upper bound in~\eqref{eq: Top dual hinge c3} can be omitted for~$K = 1.$ Since we know the form of the optimal solution~\eqref{eq: Delta optimal}, we only need to show how to compute~$\Delta_{lb},$~$\Delta_{ub}$ and~$\gamma$ for all update rules~\eqref{eq: update rules}. The following three propositions provide closed-form formulae for all three update rules. To keep the presentation as simple as possible, we postpone all proofs to Appendix~\ref{sec: toppushk family coordinate proofs}.

\begin{restatable}[Update rule~\eqref{eq: update rule a,a} for problem~\eqref{eq: Top dual hinge}]{proposition}{topruleaa}\label{prop: toppushk family hinge update a,a}
  Consider problem~\eqref{eq: Top dual hinge}, update rule~\eqref{eq: update rule a,a}, indices~$1 \leq k \leq \npos$ and~$1 \leq l \leq \npos$ and Notation~\ref{not: dual update rules}. Then the optimal solution~$\Delta^{\star}$ is given by~\eqref{eq: Delta optimal} where
  \begin{align*}
    \Delta_{lb} & = \max\{- \alphak,\; \alphal - C\}, \\
    \Delta_{ub} & = \min\{C - \alphak,\; \alphal \}, \\
    \gamma & = -\frac{s_k - s_l}{\K_{kk} + \K_{ll} - \K_{kl} - \K_{lk}}.
  \end{align*}
\end{restatable}

\begin{restatable}[Update rule~\eqref{eq: update rule a,b} for problem~\eqref{eq: Top dual hinge}]{proposition}{topruleab}\label{prop: toppushk family hinge update a,b}
  Consider problem~\eqref{eq: Top dual hinge}, update rule~\eqref{eq: update rule a,b}, indices~$1 \leq k \leq \npos$ and~$\npos + 1 \leq l \leq \ntil$ and Notation~\ref{not: dual update rules}. Let us define
  \begin{equation*}
    \beta_{\max} = \max_{j \in \{1, 2, \ldots, \ntil \} \setminus \{\hat{l}\}} \beta_j.
  \end{equation*}
  Then the optimal solution~$\Delta^{\star}$ is given by~\eqref{eq: Delta optimal} where
  \begin{align*}
    \Delta_{lb} & = 
      \begin{cases*}
        \max \Brac[c]{- \alphak, \;  -\betal} & K = 1, \\
        \max \Brac[c]{- \alphak, \;  -\betal, \; K\beta_{\max} - \sum_{i = 1}^{\npos} \alpha_i} & \textrm{otherwise},
      \end{cases*} \\
    \Delta_{ub} & = 
      \begin{cases*}
          C - \alphak & K = 1, \\
          \min \Brac[c]{C - \alphak, \; \frac{1}{K-1}\Brac{\sum_{i = 1}^{\npos} \alpha_i - K \betal}}  & \textrm{otherwise}.
      \end{cases*} \\
    \gamma & = - \frac{s_k + s_l - 1}{\K_{kk} + \K_{ll} + \K_{kl} + \K_{lk}}.
  \end{align*}
\end{restatable}

\begin{restatable}[Update rule~\eqref{eq: update rule b,b} for problem~\eqref{eq: Top dual hinge}]{proposition}{toprulebb}\label{prop: toppushk family hinge update b,b}
  Consider problem~\eqref{eq: Top dual hinge}, update rule~\eqref{eq: update rule b,b}, indices~$\npos + 1 \leq k \leq \ntil$ and~$\npos + 1 \leq l \leq \ntil$ and Notation~\ref{not: dual update rules}. Then the optimal solution~$\Delta^{\star}$ is given by~\eqref{eq: Delta optimal} where
  \begin{align*}
    \Delta_{lb} & = 
      \begin{cases*}
        - \betak & K = 1, \\
        \max \Brac[c]{- \betak,\; \betal - \frac{1}{K} \sum_{i = 1}^{\npos} \alpha_i} & \textrm{otherwise},
      \end{cases*} \\
    \Delta_{ub} & = 
      \begin{cases*}
        \betal & K = 1, \\
        \min \Brac[c]{\frac{1}{K} \sum_{i = 1}^{\npos} \alpha_i - \betak,\; \betal} & \textrm{otherwise}.
      \end{cases*} \\
    \gamma & = -\frac{s_k - s_l}{\K_{kk} + \K_{ll} - \K_{kl} - \K_{lk}}.
  \end{align*}
\end{restatable}

\subsubsection{Initialization}

For all update rules~\eqref{eq: update rules} we assumed that the current solution~$\bm{\alpha},$~$\bm{\beta}$ is feasible. So to create an iterative algorithm that solves problem~\eqref{eq: Top dual hinge} or~\eqref{eq: Top dual quadratic}, we need to have a way how to obtain an initial feasible solution. Such a task can be formally written as a projection of random variables~$\bm{\alpha}^0,$~$\bm{\beta}^0$ to the feasible set of solutions
\begin{mini}{\bm{\alpha}, \bm{\beta}}{
  \frac{1}{2} \norm{\bm{\alpha} - \bm{\alpha}^0}^2
  + \frac{1}{2} \norm{\bm{\beta} - \bm{\beta}^0}^2
  }{\label{eq: toppushk family initialization}}{}
  \addConstraint{\sum_{i = 1}^{\npos} \alpha_i}{= \sum_{j = 1}^{\ntil} \beta_j}
  \addConstraint{0 \leq \alpha_i}{\leq C, \quad i = 1, 2, \ldots, \npos,}
  \addConstraint{0 \leq \beta_j}{\leq \frac{1}{K} \sum_{i = 1}^{\npos} \alpha_i, \quad j = 1, 2, \ldots, \ntil,}
\end{mini}
where the upper bound in the second constraint depends on the used surrogate function. To solve problem~\eqref{eq: toppushk family initialization}, we follow the same approach as in~\cite{adam2020projections}. In the following theorem, we show that problem~\eqref{eq: toppushk family initialization} can be written as a system of two equations of two variables~$\lambda$ and~$\mu.$ Moreover, the theorem shows the concrete form of feasible solution~$\bm{\alpha},$~$\bm{\beta}$ that depends only on~$\lambda$ and~$\mu.$

\begin{restatable}{theorem}{topinit}\label{thm: toppushk family initialization}
  Consider problem~\eqref{eq: toppushk family initialization}, some initial solution~$\bm{\alpha}^0,$~$\bm{\beta}^0$ and denote the sorted version (in non-decreasing order) of~$\bm{\beta}^0$ as~$\bm{\beta}_{[\cdot]}^0.$ Then if the following condition holds
  \begin{equation}\label{eq:problem3_cond}
    \sum_{j = 1}^{K} \Brac{\beta_{[\ntil - K + j]}^0 + \max_{i = 1, \ldots, \npos} \alpha_i^0} \le 0,
  \end{equation}
  the optimal solution of~\eqref{eq: toppushk family initialization} amounts to~$\bm{\alpha} = \bm{\beta} = \bm{0}.$ In the opposite case, the following system of two equations
  \begin{subequations}\label{eq: toppushk family init alg}
    \begin{align}
      \sum_{i=1}^{\npos} \clip{0}{C}{ \alpha_i^0 - \lambda + \frac{1}{K} \sum_{j=1}^{\ntil} \clip[u]{0}{+\infty}{\beta_j^0 + \lambda - \mu}} - K \mu
      & = 0, \label{eq: toppushk family init alg 1} \\
      \sum_{j=1}^{\ntil} \clip{0}{\mu}{\beta_j^0 + \lambda} - K\mu
      & = 0, \label{eq: toppushk family init alg 2}
    \end{align}
  \end{subequations}
  has a solution~$(\lambda, \mu)$ with $\mu > 0,$ and the optimal solution of~\eqref{eq: toppushk family initialization} is equal to
  \begin{align*}
    \alpha_i
      & = \clip{0}{C}{\alpha_i^0 - \lambda + \frac{1}{K} \sum_{j=1}^{\ntil} \clip[u]{0}{+\infty}{\beta_j^0 + \lambda - \mu}}, \\
    \beta_j & = \clip{0}{\mu}{\beta_j^0 + \lambda}.
  \end{align*}
\end{restatable}

Theorem~\ref{thm: toppushk family initialization} shows the optimal solution of~\eqref{eq: toppushk family initialization} that depends only on~$(\lambda, \mu)$ but does not provide any way to find such a solution. In the following text, we show that the number of variables in the system of equations~\eqref{eq: toppushk family init alg} can be reduced to one. For any fixed $\mu$, we denote the function on the left-hand side of~\eqref{eq: toppushk family init alg 2} by 
\begin{equation*}
  g(\lambda; \mu) := \sum_{j=1}^{\ntil} \clip{0}{\mu}{\beta_j^0 + \lambda} - K\mu.
\end{equation*}
Then~$g$ is non-decreasing in~$\lambda$ but not necessarily strictly increasing. We denote by~$\lambda(\mu)$ any such~$\lambda$ solving~\eqref{eq: toppushk family init alg 2} for a fixed~$\mu$. Denote~$\bm{z}$ the sorted version of~$-\bm{\beta}^0$. Then we have
\begin{equation*}
  g(\lambda; \mu)
    = \sum_{\Set{j}{\lambda - z_j \in [0, \mu)}}(\lambda - z_j)
    + \sum_{\Set{j}{\lambda - z_j \ge \mu}}\mu - K\mu.
\end{equation*}
Now we can easily compute~$\lambda(\mu)$ by solving~$g(\lambda(\mu); \mu) = 0$ for fixed~$\mu.$ To get the solution efficiently, we derive Algorithm~\ref{alg: toppushk family lambda}, which can  be described as follows: Index~$i$ will run over~$\bm{z}$ while index~$j$ will run over~$\bm{z} + \mu$. At every iteration, we know the values of~$g(z_{i-1}; \mu)$ and~$g(z_{j-1}+\mu; \mu)$ and we want to evaluate~$g$ at the next point. We denote the number of indices~$j$ such that $\lambda - z_j \in[0, \mu)$ by~$d$. If~$z_i \le z_j + \mu$, then we consider~$\lambda = z_i$ and since one index enters the set~$\Set{j}{\lambda - z_j \in [0, \mu)}$, we increase~$d$ by one. On the other hand, if~$z_i > z_j + \mu$, then we consider $\lambda = z_j + \mu$ and since one index leaves the set~$\Set{j}{\lambda - z_j \in [0, \mu)}$, we decrease~$d$ by one. In both cases,~$g$ is increased by~$d$ times the difference between the new~$\lambda$ and old~$\lambda$. Once~$g$ exceeds~$0$, we stop the algorithm and linearly interpolate between the last two values. To prevent an overflow, we set~$z_{m+1} = + \infty$. Concerning the initial values, since~$z_1 \le z_1 + \mu$, we set $i=2$, $j=1$ and $d=1$. 

\begin{algorithm}
  \centering
  \begin{algorithmic}[1]
    \Require vector $-\bm{\beta}^0$ sorted into $\bm{z}$
    \State $i \gets 2$, $j \gets 1$, $d \gets 1$
    \State $\lambda \gets z_1$, $g \gets - K\mu$
    \While{$g < 0$}
      \If {$z_i \le z_j + \mu$}
        \State $g \gets g + d(z_i - \lambda)$
        \State $\lambda\gets z_i$, $d \gets d+1$, $i \gets i+1$
      \Else
        \State $g \gets g + d(z_j + \mu - \lambda)$
        \State $\lambda \gets z_j + \mu$, $d \gets d - 1$, $j \gets j + 1$
      \EndIf
    \EndWhile
    \State \textbf{return} linear interpolation of the last two values of $\lambda$
  \end{algorithmic}
  \caption{An efficient algorithm for computing~$\lambda(\mu)$ from~\eqref{eq: toppushk family initialization} for fixed~$\mu.$.}
  \label{alg: toppushk family lambda}
\end{algorithm}

Since~$\lambda(\mu)$ can be computed for fixed~$\mu$ using Algorithm~\ref{alg: toppushk family lambda}, we can define auxiliary function~$h$ in the following form
\begin{equation}\label{eq: toppushk family h}
  h(\mu)
    = \sum_{i=1}^{\npos} \clip{0}{C}{\alpha_i^0 - \lambda(\mu) + \frac{1}{K} \sum_{j=1}^{\ntil} \clip[u]{0}{+\infty}{\beta_j^0+\lambda(\mu) - \mu}} - K \mu.
\end{equation}
Then the system of equations~\eqref{eq: toppushk family init alg} is equivalent to~$h(\mu) = 0.$ The following lemma describes properties of~$h.$ Since~$h$ is decreasing in~$\mu$ on~$(0, \infty)$, any root-finding algorithm such as bisection can be used to find the optimal solution.

\begin{restatable}{lemma}{topinith}\label{lemma: toppushk family h}
  Even though~$\lambda(\mu)$ is not unique, function~$h$ from~\eqref{eq: toppushk family h} is well-defined in the sense that it gives the same value for every choice of~$\lambda(\mu)$. Moreover,~$h$ is decreasing in~$\mu$ on~$(0, + \infty)$.
\end{restatable}

\subsection{Summary}

In this section, we derived dual formulation for \TopPushK and \PatMat family of formulations. Moreover, we derived simple update rules that can be used to improve the current feasible solution. We also showed that these update rules have closed-form formulae, and therefore they are simple to compute. Finally, we showed how to find an initial feasible solution. For \TopPushK family with hinge loss, we showed the derivation in the previous section, while the derivations for \PatMat family are in Appendix~\ref{sec: Pat coordinate descent}. This section combines all these intermediate results into Algorithm~\ref{alg:Coordinate descent} and discusses its computational complexity.

\begin{algorithm*}
  \begin{minipage}{0.48\textwidth}
    \centering
    \begin{algorithmic}[1]
      \State Set~$\bm{\alpha},$~$\bm{\beta}$ using Theorem~\ref{thm: toppushk family initialization}
      \State Set~$\bm{s}$ based on~\eqref{eq: dual scores} \label{alg: line 1}
      \Repeat \label{alg: line 2}
        \State Pick random~$k$ from~$\{1, \ldots, \npos + \ntil\}$ \label{alg: line 3}
        \For{$l \in \{1, \ldots, \npos + \ntil  \}$} \label{alg: line 4}
            \State Compute~$\Delta_{l}$  \label{alg: line 5}
        \EndFor
        \State Select the best~$\Delta_{l}$ \label{alg: line 7}
        \State Update~$\bm{\alpha}$,~$\bm{\beta},$~$\bm{s}$ according to~\eqref{eq: update rules} \label{alg: line 8}
        \State \label{alg: line 9}
      \Until{stopping criterion is satisfied}
    \end{algorithmic}
  \end{minipage}%
  \hfill
  \begin{minipage}{0.48\textwidth}
    \centering
    \begin{algorithmic}[1]
      \State Set~$\bm{\alpha},$~$\bm{\beta},$~$\delta$ using Theorem~\ref{thm: patmat family initialization}
      \State Set~$\bm{s}$ based on~\eqref{eq: dual scores}
      \Repeat
        \State Pick random~$k$ from~$\{1, \ldots, \npos + \ntil \}$ 
        \For{$l \in \{1, \ldots, \npos + \ntil \}$}
            \State Compute~$\Delta_{l}$ and~$\delta_{l}$
        \EndFor
        \State Select the best~$\Delta_{l}$ and~$\delta_{l}$
        \State Update~$\bm{\alpha}$,~$\bm{\beta},$~$\bm{s}$ according to~\eqref{eq: update rules}
        \State set~$\delta \leftarrow \delta_{l}$
      \Until{stopping criterion is satisfied}
    \end{algorithmic}
  \end{minipage}
  \caption{Coordinate descent algorithm for \TopPushK family of formulations (\textbf{left}) and \PatMat  family of formulations (\textbf{right}).}
  \label{alg:Coordinate descent}
\end{algorithm*}

The left column in Algorithm~\ref{alg:Coordinate descent} describe the algorithm for \TopPushK family while the right column for \PatMat family. In step~\ref{alg: line 1} we initialize~$\bm{\alpha}$,~$\bm{\beta}$ and~$\delta$ to some feasible value using Theorem~\ref{thm: toppushk family initialization} or  Theorem~\ref{thm: patmat family initialization}. Then, based on~\eqref{eq: dual scores} we compute scores~$\bm{s}$. Each \repeatloop loop in step~\ref{alg: line 2} updates two coordinates as shown in~\eqref{eq: update rules}. In step~\ref{alg: line 3} we select a random index~$k$ and in the \forloop loop in step~\ref{alg: line 4} we compute the optimal~$(\Delta_l,\delta_l)$ for all possible combinations~$(k,l)$ as in~\eqref{eq: update rules}. In step~\ref{alg: line 7} we select the best pair~$(\Delta_l,\delta_l)$ which maximizes the coresponding objective function. Finally, based on the selected update rule we update~$\bm{\alpha}$,~$\bm{\beta}$,~$\bm{s}$ and~$\delta$ in steps~\ref{alg: line 8} and~\ref{alg: line 9}.

Now we derive the computational complexity of each \repeatloop loop from step~\ref{alg: line 2}. The computation of~$(\Delta_l,\delta_l)$ amounts to solving a quadratic optimization problem in one variable. As we showed in Sections~\ref{sec: Top coordinate descent} and~\ref{sec: Pat coordinate descent}, there is a closed-form solution and step~\ref{alg: line 5} can be performed in~$O(1)$. Since this is embedded in a \forloop loop in step~\ref{alg: line 4}, the whole complexity of this loop is~$O(\npos + \ntil)$. Step~\ref{alg: line 8} requires~$O(1)$ for the update of~$\bm{\alpha}$ and~$\bm{\beta}$ while~$O(\npos + \ntil)$ for the update of~$\bm{s}$. Since the other steps are~$O(1)$, the total complexity of the \repeatloop loop is~$O(\npos + \ntil)$. This holds only if the kernel matrix~$\K$ is precomputed. In the opposite case, all complexities must be multiplied by the cost of computation of components of~$\K$, which is~$O(d)$. This complexity analysis is summarized in Table~\ref{tab:Computational complexity}.

\begin{table}[h]
  \centering
  \begin{tabular}{lcc}
    \hline
    Operation
      & $\K$ precomputed
      & $\K$ not precomputed \\
    \hline
    Evaluation of~$\Delta_l$
      & $O(1)$
      & $O(d)$ \\
    Update of~$\bm{\alpha}$ and~$\bm{\beta}$
      & $O(1)$
      & $O(1)$ \\
    Update of~$\bm{s}$
      & $O\Brac{\npos + \ntil}$
      & $O\Brac{(\npos + \ntil)d}$ \\
    \hline
    Total per iteration
    & $O\Brac{\npos + \ntil}$
    & $O\Brac{(\npos + \ntil)d}$ \\
    \hline
  \end{tabular}
  \caption{Computational complexity of one \repeatloop loop (which updates two coordinates of~$\bm{\alpha}$ or~$\bm{\beta}$) from Algorithm~\ref{alg:Coordinate descent}.}
  \label{tab:Computational complexity}
\end{table}

% % ------------------------------------------------------------------------------
% % Numerical experiments
% % ------------------------------------------------------------------------------
\section{Numerical Experiments}\label{sec: experiments}

In this section, we describe in detail all settings used for the experiments. The section consists of five subsections. The first one discusses which formulations from Table~\ref{tab: summary formulations} we use for the experimental evaluation. In this subsection, we also introduce baseline formulations used for the comparison. In the second one, we introduce datasets used in the experiments and describe their structure. A detailed description of the datasets is then provided in separate sections with their corresponding experiment results. The third and fourth subsections contain a detailed description of performance metrics. The last subsection contains a description of tools used for implementation. All codes used for the experiments, as well as all experiment configurations, are publicly available on GitHub. We provide one respository with the code
\begin{center}
  \url{https://github.com/VaclavMacha/ClassificationAtTopDual}
\end{center}
and one repository with numerical experiments 
\begin{center}
  \url{https://github.com/VaclavMacha/ClassificationAtTopExperiments.jl}
\end{center}

\subsection{Formulations}

To simplify the setup of all experiments, we decided to focus on formulations that only use negative samples for the threshold computation, since the performance of such formulations can be compared by basic performance metrics, as shown later in Section~\ref{sec: performance criteria}. In total, we use four different formulations from Table~\ref{tab: summary formulations}, namely \TopPush, \TopPushK, \tauFPL, and \PatMatNP. Moreover, for \TopPushK, we use two different values of~$K = \{5, 10\}$ and consider the resulting formulations as separate formulations, i.e., we have \TopPushK(5) and \TopPushK(10). Similarly, for \tauFPL and \PatMatNP we use two different values of~$\tau = \{0.01, 0.05\}.$ For all formulations, we use the hinge loss defined in Notation~\ref{not: surrogates} as a surrogate function.

As a baseline formulation for comparison,  we use C-SVC variant of SVM~\cite{boser1992training, cortes1995support,chang2011libsvm} defined by
\begin{mini}{\bm{w}, b, \bm{\xi}}{
  \frac{1}{2} \norm{\bm{w}}^2 + C \sum_{i \in \I} \xi_i
  }{\label{eq: SVM}}{}
  \addConstraint{y_i}{\Brac{\bm{w}^{\top} \phi(\bm{x}_i) + b} \geq 1 - \xi_i, \quad i \in \I}
  \addConstraint{\xi_i}{\geq 0, \quad i \in \I,}
\end{mini}
where~$y_i \in \{-1, 1\}$ for all~$i \in \I$ and~$\phi(\bm{x}_i)$ maps~$\bm{x}_i$ into a higher-dimensional space (see Section~\ref{sec: kernels}). The corresponding dual form is as follows
\begin{maxi}{\bm{\alpha}}{
  - \frac{1}{2} \bm{\alpha}^{\top} \K \bm{\alpha} - \sum_{i = 1}^{\nall} \alpha_i
  }{\label{eq: SVM dual}}{}
  \addConstraint{\sum_{i = 1}^{\nall} y_i \alpha_i}{= 0}
  \addConstraint{0 \leq \alpha_i }{\leq C, \quad i = 1, 2, \ldots, \nall,}
\end{maxi}
where the kernel matrix~$\K$ is defined for all~$i, j = 1, 2, \ldots, \nall$ as
\begin{equation*}
  \K_{i,j} = y_i y_j k(\bm{x}_i, \bm{x}_j) = \phi(\bm{x}_i)^{\top} \phi(\bm{x}_j).
\end{equation*}
Note that the dual form of C-SVC is very similar to the dual forms of our formulations derived in Section~\ref{sec:Derivation of dual problems}. We will denote C-SVC as \SVM.

In total, we have five different formulations for experiments, as seen in Table~\ref{tab: formulations experiments summary}. The following section discusses which hyper-parameters are used for each formulation. Since we used a slightly different primal form (standard formulation for SVM) for the derivation of dual forms, we also show how to convert used parameters to the resulting dual forms and get identical experiment settings.

\subsection{Hyper-parameters}

The selected formulations differ in the number of available hyper-parameters. Therefore, we decided to use a fixed value for all but one of the hyper-parameters jfor each formulation. For most of the considered formulations, the only hyper-parameter is the regularization constant~$\lambda$. In our experiments, we used the following six values of this hyper-parameter
\begin{equation*}
  \lambda \in \Brac[c]{0, 10^{-5}, 10^{-4}, 10^{-3}, 10^{-2}, 10^{-1}}.
\end{equation*}
The only exceptions are the formulations derived from \PatMatNP since they also have the scaling parameter~$\vartheta.$ Since the parameter is essential for the approximation quality of the threshold, we decided to fine-tune this hyper-parameter instead of the regularization constant~$\lambda$. Therefore, we fixed~$\lambda$ to~$10^{-3}$ for \PatMatNP formulations and used the following six different values of the scaling parameter
\begin{equation*}
  \vartheta \in \Brac[c]{10^{-5}, 10^{-4}, 10^{-3}, 10^{-2}, 10^{-1}, 1}.
\end{equation*}
Since we used a slightly different (but equivalent) primal formulation for the derivation of the dual forms, we use~$\lambda$ to compute the hyper-parameter~$C$ used in these dual forms
\begin{equation*}
  C = \frac{1}{\lambda \ntil},
\end{equation*}
where~$\ntil = \nall$ for \SVM and~$\ntil = \npos$ otherwise. In all experiments, the best hyperparameter is selected based on the validation data and the appropriate performance metric. A summary of all used formulations and their hyper-parameters is in Table~\ref{tab: formulations experiments summary}.

\begin{table}[!ht]
  \centering
  \begin{tabular}{lcc}
    \hline
    \textbf{Formulation}
      & \textbf{Fixed parameters}
      & \textbf{Hyper-parameter} \\
    \hline
    \SVM
      & ---
      & $\lambda$ \\
    \TopPush
      & ---
      & $\lambda$ \\
    \TopPushK(5)
      & $K = 5$
      & $\lambda$ \\
    \TopPushK(10)
      & $K = 10$
      & $\lambda$ \\
    \tauFPL(0.01)
      & $\tau = 0.01$
      & $\lambda$ \\
    \tauFPL(0.05)
      & $\tau = 0.05$
      & $\lambda$ \\
    \PatMatNP(0.01)
      & $\tau = 0.01$
      & $\vartheta$ \\
    \PatMatNP(0.05)
      & $\tau = 0.05$
      & $\vartheta$ \\
    \hline
  \end{tabular}
  \caption{Summary of all formulations used for experiments. The first column shows the aliases used for the formulations when describing the experiment results. The second column shows fixed parameters used for each formulation, while the third column shows which hyper-parameters are tuned using the validation set.}
  \label{tab: formulations experiments summary}
\end{table}

\subsection{Datasets}

We consider various datasets summarized in Table~\ref{tab: datasets summary} for the numerical experiments. All these datasets are from the domain of image recognition. We use this domain since it is one of the most popular with plenty of publicly available datasets. MNIST~\cite{deng2012mnist} and FashionMNIST~\cite{xiao2017fashionmnist} are grayscale datasets of digits and fashion items, respectively. CIFAR100~\cite{krizhevsky2009learning} is a dataset of colored images of different items grouped into 100 classes. CIFAR10 and CIFAR20 merge these classes into 10 and 20 superclasses, respectively. Finally, SVHN2~\cite{netzer2011reading} contains colored images of house numbers. All these datasets are originally divided only into training and test sets. We select 25\% samples from the training set to obtain the validation set. Moreover, all datasets are multiclass, we need to adjust the labels to get a binary classification problem. Therefore, for each data set, we select one class as the positive class and consider the rest as the negative class.

It is worth mentioning that all datasets used in the experiments are not primarily designed for the classification at the top. We use these datasets since they are publicly available and well-known.

\begin{table}[!ht]
  \centering
  \resizebox{\columnwidth}{!}{%
    \begin{tabular}{lccrrrrrr}
      \hline
      \textbf{Dataset}
      & $y^+$
      & $d$
      & \multicolumn{2}{c}{\textbf{Train}}
      & \multicolumn{2}{c}{\textbf{Validation}}
      & \multicolumn{2}{c}{\textbf{Test}} \\
      \cline{4-9}
      &&& $n$
      & $\frac{\npos}{n}$
      & $n$
      & $\frac{\npos}{n}$
      & $n$
      & $\frac{\npos}{n}$ \\
      \hline
      MNIST
      & 1
      & $28 \times 28 \times 1$
      & 45 000
      & 11.3\%
      & 15 000
      & 11.2\%
      & 10 000
      & 11.4\% \\
      FashionMNIST
      & 1
      & $28 \times 28\times 1$
      & 45 000
      & 10.0\%
      & 15 000
      & 9.9\%
      & 10 000
      & 10.0\% \\
      CIFAR10
      & 1
      & $32 \times 32 \times 3$
      & 37 500
      & 10.0\%
      & 12 500
      & 9.9\%
      & 10 000
      & 10.0\% \\
      CIFAR20
      & 1
      & $32 \times 32 \times 3$
      & 37 500
      & 5.0\%
      & 12 500
      & 5.1\%
      & 10 000
      & 5.0\% \\
      CIFAR100
      & 1
      & $32 \times 32 \times 3$
      & 37 500
      & 1.0\%
      & 12 500
      & 1.0\%
      & 10 000
      & 1.0\% \\
      SVHN2
      & 1
      & $32 \times 32\times 3$
      & 54 944
      & 18.9\%
      & 18 313
      & 18.9\%
      & 26 032
      & 19.6\% \\
      \hline
    \end{tabular}
  }
  \caption{Structure of datasets: The training, validation and testing sets show the positive label~$y^+,$ the number of features~$d$, samples~$n$ and the fraction of positive samples~$\frac{\npos}{n}$.}
  \label{tab: datasets summary}
\end{table}

\subsection{Performance Criteria}\label{sec: performance criteria}

In this section, we describe which performance criteria are used for evaluation and how these criteria are related to the tested formulations.

As we discussed at the beginning of Section~\ref{sec: experiments}, we decided to only test formulations that minimize the false-negative rate and use only negative samples for the threshold computation. This choice allows us to use simple metrics to compare the formulations. The first metric that we use in the experiments is~$\tpratk$ defined as follows
\begin{equation*}
  \tpratk = \frac{1}{\npos} \sum_{i \in \Ipos} \Iverson{s_i \geq t} \quad \text{where} \quad t = \frac{1}{K} \sum_{j = 1}^{K} s^{-}_{[j]}.
\end{equation*}
This metric computes the true-positive rate at a threshold~$t$ defined as the mean of $K$-largest negative scores. For~$K = 1$, the threshold corresponds to the threshold used by \TopPush formulation. Otherwise, threshold~$t$ corresponds to the threshold used by \TopPushK. Moreover, since minimizing the false-negative rate is equivalent to maximizing the true-positive rate, both \TopPush and \TopPushK should optimize the $\tpratk$ metric. In the upcoming experiments, we use this metric with three different values of~$K \in \{1, 5, 10\}.$

The second metric is defined in a similar way
\begin{equation*}
  \tpratfpr = \frac{1}{\npos} \sum_{i \in \Ipos} \Iverson{s_i \geq t} \quad \text{where} \quad t
  = \max \Set{t}{\frac{1}{\nneg} \sum_{i \in \Ineg} \Iverson{s_i \geq t} \geq \tau}.
\end{equation*}
This metric computes the true-positive rate at a specific top $\tau$-quantile of negative scores. This metric is ideal for testing the performance of \tauFPL and \PatMatNP formulations since both maximize the true-positive rate and use some approximation of the true top $\tau$-quantile of negative scores as a threshold. In our experiments, we use this metric with two different values of~$\tau \in \{0.01, 0.05\}.$

The two previous metrics are specific to the formulations from our framework. However, we should also test if the baseline formulations work correctly. Since the baseline method is designed to optimize overall performance, we use the area under the ROC curve to measure the overall performance. The summary of all used metrics is in Table~\ref{tab: metrics summary}.

\begin{table}[!ht]
  \centering
  \begin{tabular}{lcccccc}
    \hline
    \textbf{Formulation}
      & $\auroc$
      & \multicolumn{3}{c}{$\tpratk$}
      & \multicolumn{2}{c}{$\tpratfpr$} \\
    \cline{3-7}
      && $1$  
      & $5$
      & $10$
      & $0.01$
      & $0.05$ \\
    \hline
    \SVM
      & \yesmark
      & ---
      & ---
      & ---
      & ---
      & --- \\
    \TopPush
      & ---
      & \yesmark
      & ---
      & ---
      & ---
      & --- \\
    \TopPushK(5)
      & ---
      & ---
      & \yesmark
      & ---
      & ---
      & --- \\
    \TopPushK(10)
      & ---
      & ---
      & ---
      & \yesmark
      & ---
      & --- \\
    \tauFPL(0.01) and \PatMatNP(0.01)
      & ---
      & ---
      & ---
      & ---
      & \yesmark
      & --- \\
    \tauFPL(0.05) and \PatMatNP(0.05)
      & ---
      & ---
      & ---
      & ---
      & ---
      & \yesmark \\
    \hline
  \end{tabular}
  \caption{The summary of all used performance metrics used for evaluation. In total, we use six different metrics and nine different formulations. For each formulation~\yesmark denotes the metric in which the formulation should be the best.}
  \label{tab: metrics summary}
\end{table}

\subsection{Critical Difference Diagrams}\label{sec: cd evaluation}

All metrics from Section~\ref{sec: performance criteria} can be used to compare different formulations on a single dataset. However, these metrics are unsuitable for comparing multiple formulations on multiple datasets. To address this issue, we use the Friedman test~\cite{friedman1940comparison} as suggested in~\cite{demvsar2006statistical}.

Consider that we have~$m$ datasets, and~$k$ formulations. Then for each dataset~$i$, each formulation~$j$ is ranked by rank~$r^i_j$ according to some performance criterium. Any performance metric from the previous section can be used. The formulation that provides the best result gets ranked 1; the second best gets ranked 2, and so on. If two formulations provide the same results, the average ranks are assigned. The average rank overall dataset for formulation~$j$ is computed as
\begin{equation*}
  R_j = \frac{1}{m} \sum_{i = 1}^{m} r^{i}_{j}.
\end{equation*}
The Friedman test compares the average ranks of formulations under the null hypothesis, which states that all formulations are equivalent. Therefore, their average ranks should be equal. If the null hypothesis is rejected, we proceed with the post hoc Nemenyi test~\cite{nemenyi1963distribution} that compares all formulations to each other. The performance of the two formulations is significantly different if the corresponding average
ranks differ by at least the critical difference
\begin{equation*}
  CD = q_{\alpha} \sqrt{\frac{k(k + 1)}{6m}},
\end{equation*}
where critical values~$q_{\alpha}$ are based on the Studentized range statistic divided by~$\sqrt{2},$ see Table 5(a) in~\cite{demvsar2006statistical}. The results of this post hoc test can be easily visualized using critical difference diagrams proposed in~\cite{demvsar2006statistical}. The $x$-axis of such a diagram shows the average rank over all datasets for each formulation. Formulations that are not significantly different according to the Nemenyi test are connected using a green horizontal line. As an example, see Figure~\ref{fig: dual gauss CD}.

\subsection{Implementation}

For the implementation of all experiments, we use the Julia programming language~\cite{bezanson2017julia}. All formulations are implemented from scratch. Only for SVM, we use the Julia wrapper for the LIBSVM library~\cite{chang2011libsvm}.

\subsection{Results}\label{sec: results dual}

In this section, we present results for a dual form of formulations from Table~\ref{tab: formulations experiments summary} with a Gaussian kernel model. For training, we use the coordinate descent algorithm introduced in Section~\ref{sec: coordinate descent}. We set a number of steps to 20 epochs. For all experiments, we use precomputed kernel matrix with a Gaussian kernel function defined as 
\begin{equation*}
  k(\bm{x}_i, \bm{x}_j) = \exp\Brac[c]{- \frac{\norm{\bm{x}_i - \bm{x}_j}^2}{d}},
\end{equation*}
where~$d$ is the dimension of the primal problem. We used this value of~$d$ since it is the default setting for the Gaussian kernel function in LIBSVM~\cite{chang2011libsvm}. We only use one kernel function for computational reasons. In addition, we are more interested in the comparison of methods between each other than in obtaining the best results possible.

In Figure~\ref{fig: dual convergence}, we investigate the convergence of the coordinate descend algorithm introduced in Section~\ref{sec: coordinate descent} for three formulations, namely \TopPush, \TopPushK, and \PatMatNP. In each column, we show the primal and dual objective function convergence for one formulation. To solve the primal problem, we used full gradient descent. Computation of the full gradient is computationally intensive, even for relatively small datasets such as MNIST. Therefore, for this experiment (and only for this experiment) we use the Ionosphere dataset~\cite{sigillito1989classification}, which is small. We can see that \TopPush and \TopPushK converge to the same objective for primal and dual problems. It means that both problems were solved to optimality. However, there is a little gap between the optimal primal and dual form solution for \PatMatNP. In other words, \PatMatNP may suffer from convergence issues when solving the proposed coordinate descent algorithm.

For comparison we use critical difference diagrams introduced in Section~\ref{sec: cd evaluation}. One of the basic assumptions of the critical difference diagrams to work appropriately is a large number of used datasets. Since we performed all experiments for each formulation and each dataset ten times with different random seeds for train/valid/test split, we decided to consider each of these runs as a separate dataset. It is important to say that we use this setting only for the critical difference diagrams. Since the critical diagrams show the relative performance of the formulations against each other, we can easily see if any formulation is significantly worse or better. However, the critical diagrams do not provide any information on the actual performance of the formulations. Therefore, even if one formulation outperforms other tested formulations, it does not mean that its performance is good.

To address the issue above, we also compare concrete performance metrics on each dataset separately. Since we have six hyper-parameters for each formulation, we always select the best result for each formulation on the validation set based on the criterion for which the specific formulation is optimized. Then for each formulation, we select the median of the best results from ten independent runs. Moreover, the best result for each dataset is highlighted in green, while the worst result is highlighted in red.

From Figure~\ref{fig: dual gauss CD} and Table~\ref{tab: dual gauss medians}, we make several observations:
\begin{itemize}
  \item We observe that some formulations have problems with convergence and, in some cases, even diverge for some datasets. The improper choice of the kernel function parameters can be the cause. As a result, CD diagrams may provide unreliable results. If the formulation diverges in some experiments, it immediately obtains very high ranks for these experiments that skew the final diagram. It is especially evident for \PatMatNP and \SVM formulations.
  \item Figure~\ref{fig: dual gauss CD} shows that \PatMatNP formulations provide the worst results for all metrics. It can be caused by the bad convergence of the coordinate descent algorithm, as shown in Figure~\ref{fig: dual convergence}. However, it is important to say that Figure~\ref{fig: dual gauss CD} shows only relative results. From Table~\ref{tab: dual gauss medians} is clear that even though \PatMatNP usually provides worse results than other formulations, the results are, in many cases, only slightly worse.
  \item Similarly to \PatMatNP, the \SVM formulation does not perform well for most metrics. However, as shown in Table~\ref{tab: dual gauss medians}, the results are usually only slightly worse than those of other formulations.
  \item Most formulations perform well on the criteria for which they are optimized. The only exceptions are \SVM and \PatMat formulations.
  \item Most formulations provide an $\auroc$ greater than 99\% on the MNIST and FashionMNIST datasets. These two datasets are very easy when a non-linear model is used.
  \item \tauFPL formulations work very well for $\tpratfpr = 0.01,$ $\tpratfpr = 0.05$ and $\auroc$ metric.
  \item \TopPush, \TopPushK(5) and \TopPushK(10) provides very good results for $\tpratk = 1,$ $\tpratk = 5$ and $\tpratk = 10.$
\end{itemize}

\begin{figure}
  \centering
  \begin{tikzpicture}
    \begin{groupplot}[
      group style = {
        group size = 3 by 1,
        horizontal sep = 10pt
      },
      width = \textwidth*0.35,
      grid = major,
      grid style = {dashed, gray!50, very thin},
      enlargelimits = false,
      yticklabels = {},
      xmin = 0,
      xmax = 0.4,
      ymin = 2,
      ymax = 8,
      xtick = {0, 0.1, 0.2, 0.3},
      xticklabels = {0, 0.1, 0.2, 0.3},
    ]
    \nextgroupplot[
        title = {\TopPush},
        xlabel={$t \; [s]$},
        ylabel = {Objective},
    ]
      \addplot [lineprimal] table[x index=0, y index=1] {\ConvergenceTopPush};
      \addplot [linedual]   table[x index=2, y index=3] {\ConvergenceTopPush};
    \nextgroupplot[
        title = {\TopPushK},
        xlabel={$t \; [s]$},
    ]
      \addplot [lineprimal] table[x index=0, y index=1] {\ConvergenceTopPushK};
      \addplot [linedual]   table[x index=2, y index=3] {\ConvergenceTopPushK};
    \nextgroupplot[
        title = {\PatMatNP},
        xlabel={$t \; [s]$},
        xmin = 0,
        xmax = 1,
        xtick = {0, 0.25, 0.5, 0.75},
        xticklabels = {0, 0.25, 0.5, 0.75},
        ymin = 5,
        ymax = 20
    ]
      \addplot [lineprimal] table[x index=0, y index=1] {\ConvergencePatMat};
      \addplot [linedual]   table[x index=2, y index=3] {\ConvergencePatMat};
    \end{groupplot}
  \end{tikzpicture}
  \caption{Convergence of the objectives for the primal (red line) and dual (blue dashed line) forms with linear kernel.}
  \label{fig: dual convergence}
\end{figure}

\begin{figure}[!p]
  \centering
  \begin{tikzpicture}  \node at (4.5,2.4) {\underline{$\auroc$}}; 
    \draw (1,0) -- (8,0); 
    \foreach \x in {1,...,8} \draw (\x,0.1) -- (\x,-0.1) node[anchor=north]{$\x$}; 
    \draw[line_node] (2.14,0) -- (2.14,0.4) -- (0.9, 0.4) node[anchor=east] {\tauFPL(0.05)}; 
    \draw[line_node] (3.22,0) -- (3.22,0.8) -- (0.9, 0.8) node[anchor=east] {\tauFPL(0.01)}; 
    \draw[line_node] (4.24,0) -- (4.24,1.2) -- (0.9, 1.2) node[anchor=east] {\TopPushK(10)}; 
    \draw[line_node] (4.58,0) -- (4.58,1.6) -- (0.9, 1.6) node[anchor=east] {\TopPushK(5)}; 
    \draw[line_node] (4.77,0) -- (4.77,1.6) -- (8.1, 1.6) node[anchor=west] {\SVM}; 
    \draw[line_node] (5.22,0) -- (5.22,1.2) -- (8.1, 1.2) node[anchor=west] {\TopPush}; 
    \draw[line_node] (5.91,0) -- (5.91,0.8) -- (8.1, 0.8) node[anchor=west] {\PatMatNP(0.01)}; 
    \draw[line_node] (5.92,0) -- (5.92,0.4) -- (8.1, 0.4) node[anchor=west] {\PatMatNP(0.05)}; 
    \draw[line_cv] (2.14,0.2) -- (3.22, 0.2); 
    \draw[line_cv] (3.22,0.4) -- (4.58, 0.4); 
    \draw[line_cv] (4.24,0.2) -- (5.22, 0.2); 
    \draw[line_cv] (4.58,0.2) -- (5.92, 0.2);

    \node at (4.5,5.9) {\underline{$\tpratfpr = 0.05$}}; 
    \draw (1,3.5) -- (8,3.5); 
    \foreach \x in {1,...,8} \draw (\x,3.6) -- (\x,3.4) node[anchor=north]{$\x$}; 
    \draw[line_node] (2.18,3.5) -- (2.18,3.9) -- (0.9, 3.9) node[anchor=east] {\tauFPL(0.05)}; 
    \draw[line_node] (3.86,3.5) -- (3.86,4.3) -- (0.9, 4.3) node[anchor=east] {\tauFPL(0.01)}; 
    \draw[line_node] (4.5,3.5) -- (4.5,4.7) -- (0.9, 4.7) node[anchor=east] {\TopPushK(10)}; 
    \draw[line_node] (4.73,3.5) -- (4.73,5.1) -- (0.9, 5.1) node[anchor=east] {\SVM}; 
    \draw[line_node] (4.88,3.5) -- (4.88,5.1) -- (8.1, 5.1) node[anchor=west] {\TopPush}; 
    \draw[line_node] (4.92,3.5) -- (4.92,4.7) -- (8.1, 4.7) node[anchor=west] {\TopPushK(5)}; 
    \draw[line_node] (5.41,3.5) -- (5.41,4.3) -- (8.1, 4.3) node[anchor=west] {\PatMatNP(0.01)}; 
    \draw[line_node] (5.52,3.5) -- (5.52,3.9) -- (8.1, 3.9) node[anchor=west] {\PatMatNP(0.05)}; 
    \draw[line_cv] (3.86,3.7) -- (4.92, 3.7); 
    \draw[line_cv] (4.5,3.9) -- (5.52, 3.9);

    \node at (4.5,9.4) {\underline{$\tpratfpr = 0.01$}}; 
    \draw (1,7.0) -- (8,7.0); 
    \foreach \x in {1,...,8} \draw (\x,7.1) -- (\x,6.9) node[anchor=north]{$\x$}; 
    \draw[line_node] (2.53,7.0) -- (2.53,7.4) -- (0.9, 7.4) node[anchor=east] {\tauFPL(0.05)}; 
    \draw[line_node] (3.12,7.0) -- (3.12,7.8) -- (0.9, 7.8) node[anchor=east] {\tauFPL(0.01)}; 
    \draw[line_node] (4.16,7.0) -- (4.16,8.2) -- (0.9, 8.2) node[anchor=east] {\SVM}; 
    \draw[line_node] (4.47,7.0) -- (4.47,8.6) -- (0.9, 8.6) node[anchor=east] {\TopPushK(10)}; 
    \draw[line_node] (4.75,7.0) -- (4.75,8.6) -- (8.1, 8.6) node[anchor=west] {\TopPush}; 
    \draw[line_node] (5.04,7.0) -- (5.04,8.2) -- (8.1, 8.2) node[anchor=west] {\TopPushK(5)}; 
    \draw[line_node] (5.89,7.0) -- (5.89,7.8) -- (8.1, 7.8) node[anchor=west] {\PatMatNP(0.01)}; 
    \draw[line_node] (6.04,7.0) -- (6.04,7.4) -- (8.1, 7.4) node[anchor=west] {\PatMatNP(0.05)}; 
    \draw[line_cv] (2.53,7.2) -- (3.12, 7.2); 
    \draw[line_cv] (3.12,7.4) -- (4.47, 7.4); 
    \draw[line_cv] (4.16,7.2) -- (5.04, 7.2); 
    \draw[line_cv] (4.75,7.2) -- (6.04, 7.2);

    \node at (4.5,12.9) {\underline{$\tpratk =10$}}; 
    \draw (1,10.5) -- (8,10.5); 
    \foreach \x in {1,...,8} \draw (\x,10.6) -- (\x,10.4) node[anchor=north]{$\x$}; 
    \draw[line_node] (2.7,10.5) -- (2.7,10.9) -- (0.9, 10.9) node[anchor=east] {\tauFPL(0.01)}; 
    \draw[line_node] (2.96,10.5) -- (2.96,11.3) -- (0.9, 11.3) node[anchor=east] {\TopPushK(10)}; 
    \draw[line_node] (3.13,10.5) -- (3.13,11.7) -- (0.9, 11.7) node[anchor=east] {\TopPushK(5)}; 
    \draw[line_node] (3.3,10.5) -- (3.3,12.1) -- (0.9, 12.1) node[anchor=east] {\TopPush}; 
    \draw[line_node] (4.48,10.5) -- (4.48,12.1) -- (8.1, 12.1) node[anchor=west] {\SVM}; 
    \draw[line_node] (5.06,10.5) -- (5.06,11.7) -- (8.1, 11.7) node[anchor=west] {\tauFPL(0.05)}; 
    \draw[line_node] (7.11,10.5) -- (7.11,11.3) -- (8.1, 11.3) node[anchor=west] {\PatMatNP(0.01)}; 
    \draw[line_node] (7.26,10.5) -- (7.26,10.9) -- (8.1, 10.9) node[anchor=west] {\PatMatNP(0.05)}; 
    \draw[line_cv] (2.7,10.7) -- (3.3, 10.7); 
    \draw[line_cv] (3.13,10.9) -- (4.48, 10.9); 
    \draw[line_cv] (4.48,10.7) -- (5.06, 10.7); 
    \draw[line_cv] (7.11,10.7) -- (7.26, 10.7);

    \node at (4.5,16.4) {\underline{$\tpratk =5$}}; 
    \draw (1,14.0) -- (8,14.0); 
    \foreach \x in {1,...,8} \draw (\x,14.1) -- (\x,13.9) node[anchor=north]{$\x$}; 
    \draw[line_node] (2.82,14.0) -- (2.82,14.4) -- (0.9, 14.4) node[anchor=east] {\TopPushK(10)}; 
    \draw[line_node] (3.01,14.0) -- (3.01,14.8) -- (0.9, 14.8) node[anchor=east] {\tauFPL(0.01)}; 
    \draw[line_node] (3.05,14.0) -- (3.05,15.2) -- (0.9, 15.2) node[anchor=east] {\TopPush}; 
    \draw[line_node] (3.06,14.0) -- (3.06,15.6) -- (0.9, 15.6) node[anchor=east] {\TopPushK(5)}; 
    \draw[line_node] (4.31,14.0) -- (4.31,15.6) -- (8.1, 15.6) node[anchor=west] {\SVM}; 
    \draw[line_node] (5.87,14.0) -- (5.87,15.2) -- (8.1, 15.2) node[anchor=west] {\tauFPL(0.05)}; 
    \draw[line_node] (6.8,14.0) -- (6.8,14.8) -- (8.1, 14.8) node[anchor=west] {\PatMatNP(0.01)}; 
    \draw[line_node] (7.08,14.0) -- (7.08,14.4) -- (8.1, 14.4) node[anchor=west] {\PatMatNP(0.05)}; 
    \draw[line_cv] (2.82,14.2) -- (3.06, 14.2); 
    \draw[line_cv] (3.01,14.4) -- (4.31, 14.4); 
    \draw[line_cv] (5.87,14.2) -- (7.08, 14.2);

    \node at (4.5,19.9) {\underline{$\tpratk =1$}}; 
    \draw (1,17.5) -- (8,17.5); 
    \foreach \x in {1,...,8} \draw (\x,17.61) -- (\x,17.39) node[anchor=north]{$\x$}; 
    \draw[line_node] (2.78,17.5) -- (2.78,17.9) -- (0.9, 17.9) node[anchor=east] {\TopPush}; 
    \draw[line_node] (2.98,17.5) -- (2.98,18.3) -- (0.9, 18.3) node[anchor=east] {\TopPushK(5)}; 
    \draw[line_node] (3.08,17.5) -- (3.08,18.7) -- (0.9, 18.7) node[anchor=east] {\TopPushK(10)}; 
    \draw[line_node] (3.59,17.5) -- (3.59,19.1) -- (0.9, 19.1) node[anchor=east] {\tauFPL(0.01)}; 
    \draw[line_node] (5.42,17.5) -- (5.42,19.1) -- (8.1, 19.1) node[anchor=west] {\SVM}; 
    \draw[line_node] (5.74,17.5) -- (5.74,18.7) -- (8.1, 18.7) node[anchor=west] {\tauFPL(0.05)}; 
    \draw[line_node] (6.02,17.5) -- (6.02,18.3) -- (8.1, 18.3) node[anchor=west] {\PatMatNP(0.01)}; 
    \draw[line_node] (6.37,17.5) -- (6.37,17.9) -- (8.1, 17.9) node[anchor=west] {\PatMatNP(0.05)}; 
    \draw[line_cv] (2.78,17.7) -- (3.59, 17.7); 
    \draw[line_cv] (5.42,17.7) -- (6.37, 17.7); 
  \end{tikzpicture}
  \caption{\textbf{Dual formulations with a gaussian kernel:} Critical difference diagrams (level of importance 0.05) of the Nemenyi post hoc test for the Friedman test. Each diagram shows the mean rank of each method, with rank one being the best. The green horizontal lines group methods with mean ranks that are not significantly different. The critical difference diagrams were computed for mean rank averages over all datasets.}
  \label{fig: dual gauss CD}
\end{figure}

\begin{table}[!p]
  \centering
  \underline{$\tpratk =10$}
  \vspace{0.25cm}\\
  \resizebox{\columnwidth}{!}{% 
    \begin{tabular}{lcccccc}
      \hline
      \textbf{Formulation}
        & \textbf{MNIST}
        & \textbf{FashionMNIST}
        & \textbf{CIFAR10}
        & \textbf{CIFAR20}
        & \textbf{CIFAR100}
        & \textbf{SVHN2}\\
      \hline
      \SVM
        & 97.89
        & \best{95.40}
        & 9.10
        & 4.90
        & \best{11.50}
        & 4.52 \\
      \TopPush
        & 97.62
        & 94.80
        & 10.45
        & \best{6.10}
        & 11.00
        & 5.23 \\
      \TopPushK(5)
        & 97.97
        & 94.90
        & 10.05
        & 6.00
        & 11.0
        & 5.07 \\
      \TopPushK(10)
        & 97.97
        & 94.90
        & 9.85
        & \best{6.10}
        & 11.00
        & 5.18 \\
      \tauFPL(0.01)
        & \best{98.02}
        & 95.05
        & \best{10.70}
        & 5.90
        & 10.5
        & \best{5.25} \\
      \tauFPL(0.05)
        & 92.56
        & \worst{92.20}
        & 10.15
        & 5.10
        & 10.0
        & 5.24 \\
      \PatMatNP(0.01)
        & 88.37
        & 92.50
        & \worst{7.45}
        & 1.40
        & \worst{5.00}
        & \worst{4.02} \\
      \PatMatNP(0.05)
        & \worst{52.60}
        & 92.50
        & \worst{7.45}
        & \worst{1.30}
        & \worst{5.00}
        & 4.05 \\
      \hline
    \end{tabular}
  }
  \vspace{0.25cm}\\
  \underline{$\tpratfpr = 0.05$}
  \vspace{0.25cm}\\
  \resizebox{\columnwidth}{!}{% 
    \begin{tabular}{lcccccc}
      \hline
      \textbf{Formulation}
        & \textbf{MNIST}
        & \textbf{FashionMNIST}
        & \textbf{CIFAR10}
        & \textbf{CIFAR20}
        & \textbf{CIFAR100}
        & \textbf{SVHN2}\\
      \hline
      \SVM
        & 99.74
        & 98.90
        & 60.00
        & \best{44.80}
        & 59.00
        & \worst{59.72} \\
      \TopPush
        & 99.74
        & 98.80
        & 57.10
        & \worst{37.70}
        & 59.50
        & 72.54 \\
      \TopPushK(5)
        & \best{99.82}
        & 98.90
        & 56.25
        & 38.80
        & \worst{57.50}
        & 71.40 \\
      \TopPushK(10)
        & \best{99.82}
        & 98.90
        & 56.90
        & 38.70
        & 58.00
        & 71.61 \\
      \tauFPL(0.01)
        & \best{99.82}
        & 98.90
        & 58.10
        & 39.10
        & 59.00
        & 73.52 \\
      \tauFPL(0.05)
        & 99.74
        & \best{99.10}
        & \best{60.80}
        & 44.40
        & 61.00
        & \best{74.26} \\
      \PatMatNP(0.01)
        & \worst{99.30}
        & \worst{98.10}
        & \worst{54.70}
        & 44.60
        & 62.50
        & 63.47 \\
      \PatMatNP(0.05)
        & 99.38
        & \worst{98.10}
        & \worst{54.70}
        & 44.50
        & \best{63.50}
        & 63.48 \\
      \hline
    \end{tabular}
  }
  \vspace{0.25cm}\\
  \underline{$\auroc$}
  \vspace{0.25cm}\\
  \resizebox{\columnwidth}{!}{% 
    \begin{tabular}{lcccccc}
      \hline
      \textbf{Formulation}
        & \textbf{MNIST}
        & \textbf{FashionMNIST}
        & \textbf{CIFAR10}
        & \textbf{CIFAR20}
        & \textbf{CIFAR100}
        & \textbf{SVHN2}\\
      \hline
      \SVM
        & 99.94
        & 99.66
        & 90.02
        & 79.75
        & 87.80
        & \worst{90.14} \\
      \TopPush
        & 99.94
        & 99.56
        & 89.35
        & 79.06
        & \worst{87.03}
        & 92.77 \\
      \TopPushK(5)
        & 99.95
        & 99.64
        & 89.05
        & 79.13
        & 87.21
        & 92.60 \\
      \TopPushK(10)
        & 99.95
        & 99.67
        & 89.16
        & 79.27
        & 87.78
        & 92.67 \\
      \tauFPL(0.01)
        & \best{99.97}
        & 99.68
        & 89.83
        & 79.07
        & 87.64
        & 92.98 \\
      \tauFPL(0.05)
        & 99.93
        & \best{99.80}
        & \best{90.34}
        & \best{80.17}
        & 88.56
        & \best{93.16} \\
      \PatMatNP(0.01)
        & \worst{99.78}
        & \worst{99.40}
        & 87.62
        & 78.82
        & \best{89.78}
        & 90.80 \\
      \PatMatNP(0.05)
        & \worst{99.78}
        & \worst{99.40}
        & \worst{87.61}
        & \worst{78.76}
        & 89.52
        & 90.82 \\
      \hline
    \end{tabular}
  }
  \caption{\textbf{Dual formulations with a gaussian kernel:} Each table corresponds to one performance metric, and all presented results are medians of ten independent runs for each dataset and formulation pair. The best result for each dataset is highlighted in green, while the worst result is highlighted in red. For better readability, we have reduced the number of discussed metrics compared to Figure~\ref{fig: dual gauss CD}.}
  \label{tab: dual gauss medians}
\end{table}

% ------------------------------------------------------------------------------
% Conclusion
% ------------------------------------------------------------------------------
\section{Conclusion}\label{sec:Conclusion}

In this paper, we analyzed and extended the general framework for binary classification on top samples from~\cite{adam2021general} to nonlinear problems. Achieved results can be summarized as follows:
\begin{itemize}
    \item We showed that all presented formulations (except for \Grill and \GrillNP) can be divided into two families based on the form of the constraints, namely \TopPushK and \PatMat family of formulations. We derived dual forms for \TopPushK and \PatMat family of formulations. Moreover, for both these formulations we show how to incorporate non-linear kernels.
    \item We proposed a new coordinate descent algorithm for solving dual forms of \TopPushK and \PatMat family of formulations. The resulting algorithm depends on the used surrogate function. Therefore, we derived the closed-form formulae for selected surrogate functions. Since the algorithm needs a feasible solution for initialization, we also showed how to find such a solution.
    \item We performed a numerical analysis of the proposed method.
\end{itemize}

% ------------------------------------------------------------------------------
% Bibliography
% ------------------------------------------------------------------------------
\printbibliography[
  heading=bibintoc,
  title={Bibliography}
]

% ------------------------------------------------------------------------------
% Appendix
% ------------------------------------------------------------------------------
\appendix

\section{Derivation of Dual Problems}

\subsection{Family of \TopPushK Formulations}

\topdual*

\begin{proof}
  We show the proof only for \TopPushK formulation, i.e., the decision threshold is computed only from negative samples. The proof for the remaining formulations is identical. Firstly, we derive an alternative formulation to formulation~\eqref{eq: toppushk family}. Using Lemma~1 from~\cite{ogryczak2003minimizing}, we can rewrite the formula for the decision threshold to the following form
  \begin{equation*}
    \sum_{j = 1}^{K} s^-_{[j]} = \min_{t} \Brac[c]{Kt + \sum_{j = 1}^{\nneg} \max\{0, \; s^-_j - t\}}.
  \end{equation*}
  By substituting this formula into the objective function of~\eqref{eq: toppushk family}, we get
  \begin{align*}
    \sum_{i = 1}^{\npos} l\Brac{\frac{1}{K}\sum_{j = 1}^{K} s^-_{[j]} - s^+_{i}}
      & = \sum_{i = 1}^{\npos} l\Brac{ \frac{1}{K} \min_{t} \Brac[c]{Kt + \sum_{j = 1}^{\nneg} \max\Brac[c]{0, \; s^-_j - t}} - s^+_{i}} \\
      & = \min_{t} \; \sum_{i = 1}^{\npos} l\Brac{t + \frac{1}{K} \sum_{j = 1}^{\nneg} \max\Brac[c]{0, \; s^-_j - t} - s^+_{i}}.
  \end{align*}
  where the last equality follows from the fact that the surrogate function~$l$ is non-decreasing. The max operator can be replaced using an auxiliary variable~$\bm{z} \in \R^{\nneg}$ that fulfills~$z _j \geq s^-_j - t$ and~$z _j \geq 0$ for all~$j = 1, \ldots, \; \nneg.$ Furthermore, we use auxilliary variable~$\bm{y} \in \R^{\npos}$ defined for all~$i = 1, \ldots, \; \npos$ as
  \begin{equation*}
    y_i = t + \frac{1}{K} \sum_{j = 1}^{\nneg} z_j - s^+_i.
  \end{equation*}
  The combination of all the above relations and the use of a linear model yields to
  \begin{mini*}{\bm{w}, t, \bm{y}, \bm{z}}{
    \frac{1}{2} \norm{\bm{w}}_{2}^{2}+ C \sum_{i = 1}^{\npos} l(y_i)
    }{}{}
    \addConstraint{y_i}{= t + \frac{1}{K} \sum_{j = 1}^{\nneg} z_j - \bm{w}^{\top} \bm{x}^+_i, \quad}{i = 1, \; 2, \ldots, \; \npos}
    \addConstraint{z_j}{\geq \bm{w}^{\top} \bm{x}^-_j - t,}{j = 1, \; 2, \ldots, \; \nneg}
    \addConstraint{z_j}{\geq 0,}{j = 1, \; 2, \ldots, \; \nneg,}
  \end{mini*}

  The Lagrangian of this formulation is defined as 
  \begin{align*}
    \mathcal{L}(\bm{w}, t, \bm{y}, \bm{z}; \bm{\alpha}, \bm{\beta}, \bm{\gamma})
     & = \frac{1}{2} \norm{\bm{w}}_{2}^{2}
       + C \sum_{i = 1}^{\npos} l(y_i)
       + \sum_{i = 1}^{\npos} \alpha_i \Brac{t + \frac{1}{K} \sum_{j = 1}^{\nneg} z_j - \bm{w}^{\top} \bm{x}^+_i - y_i} \\
     & + \sum_{j = 1}^{\nneg} \beta_j \Brac{\bm{w}^{\top} \bm{x}^-_j - t - z_j}
       - \sum_{j = 1}^{\nneg} \gamma_j z_j,
  \end{align*}
  with feasibility conditions~$\beta_j \geq 0$ and~$\gamma_j \geq 0$ for all~$j = 1, \ldots, \; \nneg.$ Since the Lagrangian~$\mathcal{L}$ is separable in primal variables, it can be minimized with respect to each variable separately. Then the dual objective function (to be maximized) reads
  \begin{subequations}\label{eq: TopPushK dual function}
    \begin{align}
      g(\bm{\alpha}, \bm{\beta}, \bm{\gamma})
        & = \min_{\bm{w}} \; \frac{1}{2} \norm{\bm{w}}_{2}^{2}
          - \bm{w}^{\top} \Brac{\sum_{i = 1}^{\npos} \alpha_i \bm{x}^+_i - \sum_{j = 1}^{\nneg} \beta_j \bm{x}^-_j} \label{eq: TopPushK dual function w}\\
        & + \min_{t} \; t \Brac{\sum_{i = 1}^{\npos} \alpha_i - \sum_{j = 1}^{\nneg} \beta_j} \label{eq: TopPushK dual function t}\\
        & + \min_{\bm{y}} \; C \sum_{i = 1}^{\npos} \Brac{l(y_i) - \frac{\alpha_i}{C}y_i} \label{eq: TopPushK dual function y}\\
        & + \min_{\bm{z}} \; \sum_{j = 1}^{\nneg} \Brac{\frac{1}{K} \sum_{i = 1}^{\npos} \alpha_i - \beta_j - \gamma_j}z_j \label{eq: TopPushK dual function z}
    \end{align}
  \end{subequations}
  
  From optimality conditions with respect to~$\bm{w},$ we deduce 
  \begin{equation*}
    \bm{w}
      = \sum_{i = 1}^{\npos} \alpha_i \bm{x}^+_i - \sum_{j = 1}^{\nneg} \beta_j \bm{x}^-_j
      = \Matrix{\X^+ \\ - \X^-}^\top \vecab
    \; \implies \;
    \frac{1}{2} \norm{\bm{w}}_{2}^{2} - \bm{w}^{\top} \Brac{\sum_{i = 1}^{\npos} \alpha_i \bm{x}^+_i - \sum_{j = 1}^{\nneg} \beta_j \bm{x}^-_j}
      = - \frac{1}{2} \vecab^{\top} \Kneg \vecab,
  \end{equation*}
  where we use Notation~\ref{not: kernel matrix}. It mean, that we get the first part of the objective function~\eqref{eq: toppushk family dual L}, ane we also get the relation~\eqref{eq: toppushk family dual to primal} between primal and dual variables.
  
  Optimality condition with respect to~$t$ reads 
  \begin{equation*}
    \sum_{i = 1}^{\npos} \alpha_i - \sum_{j = 1}^{\nneg} \beta_j = 0,
  \end{equation*}
  and implies constrain~\eqref{eq: toppushk family dual c1}.
  
  Similarly, optimality condition of~\eqref{eq: TopPushK dual function z} with respect to~$\bm{z}$ reads for all $j = 1, \ldots, \; \nneg$ as 
  \begin{equation*}
    \frac{1}{K} \sum_{i = 1}^{\npos} \alpha_i - \beta_j - \gamma_j = 0.
  \end{equation*}
  Plugging the feasibility condition~$\gamma_j \geq 0$ into this equality and combining it with the feasibility conditions~$\beta_j \geq 0,$ yields constraint~\eqref{eq: toppushk family dual c2}.
  
  Finally, the second part of the objective function~\eqref{eq: toppushk family dual L} follows from Definition~\ref{def: conjugate} of the conjugate function. Using the definition, minimization of~\eqref{eq: TopPushK dual function y} with respect to~$\bm{y}$ yields
  \begin{equation*}
    C \min_{y_i} \Brac{l(y_i) - \frac{\alpha_i}{C} y_i} = - C l^{\star} \Brac{\frac{\alpha_i}{C}},
  \end{equation*}
  for all $i = 1, \ldots, \; \npos,$ which finishes the proof for \TopPushK. For \TopPush, we have~$K = 1.$ From~\eqref{eq: toppushk family dual c1} and non-negativity of~$\beta_j$ we deduce that the upper bound in~\eqref{eq: toppushk family dual c2} is always fulfilled and can be omitted.
\end{proof}

\subsection{Family of \PatMat Formulations}

\patdual*

\begin{proof}
  For simplicity, we show the proof only for \PatMatNP, i.e. the threshold is computed only from negative samples. Let us first realize that formulation~\eqref{eq: patmat family} is equivalent to the following formulation
  \begin{mini*}{\bm{w}, t, \bm{y}, \bm{z}}{
    \frac{1}{2} \norm{\bm{w}}_{2}^{2}+ C \sum_{i = 1}^{\npos} l(y_i)
    }{}{}
    \addConstraint{}{\sum_{j = 1}^{\nneg} l(\vartheta z_i)\leq \nneg \tau}
    \addConstraint{}{y_i = t - \bm{w}^{\top} \bm{x}^+_i, \quad}{i = 1, \; 2, \ldots, \; \npos}
    \addConstraint{}{z_j = \bm{w}^{\top} \bm{x}^-_j - t, \quad}{j = 1, \; 2, \ldots, \; \nneg.}
  \end{mini*}

  The corresponding Lagrangian then reads
  \begin{align*}
    \mathcal{L}(\bm{w}, t, \bm{y}, \bm{z}; \bm{\alpha}, \bm{\beta}, \delta)
    & = \frac{1}{2} \norm{\bm{w}}_{2}^{2}
      + C \sum_{i = 1}^{\npos} l(y_i)
      + \sum_{i = 1}^{\npos} \alpha_i (t - \bm{w}^{\top}\bm{x}^+_{i} - y_i) \\
    & + \sum_{j = 1}^{\nneg} \beta_j(\bm{w}^{\top}\bm{x}^-_j - t - z_j)
      + \delta \Brac{\sum_{j = 1}^{\nneg} l(\vartheta z_j) - \nneg \tau}.
  \end{align*}
  with feasibility condition~$\delta \geq 0.$ Since the Lagrangian~$\mathcal{L}$ is separable in primal variables, it can be minimized with respect to each variable separately. Then the dual objective function (to be maximized) can be rewritten as follows
  \begin{subequations}\label{eq: PatMat dual function}
    \begin{align}
      g(\bm{\alpha}, \bm{\beta}, \delta)
        & = \min_{\bm{w}} \; \frac{1}{2} \norm{\bm{w}}_{2}^{2}
          - \bm{w}^{\top} \Brac{\sum_{i = 1}^{\npos} \alpha_i \bm{x}^+_i - \sum_{j = 1}^{\nneg} \beta_j \bm{x}^-_j} \label{eq: PatMat dual function w}\\
        & + \min_{t} \; t \Brac{\sum_{i = 1}^{\npos} \alpha_i - \sum_{j = 1}^{\nneg} \beta_j} \label{eq: PatMat dual function t} \\
        & + \min_{\bm{y}} \; C \sum_{i = 1}^{\npos} \Brac{l(y_i) - \frac{\alpha_i}{C}y_i} \label{eq: PatMat dual function y} \\
        & + \min_{\bm{z}} \; \delta \sum_{j = 1}^{\nneg} \Brac{l(\vartheta z_j) - \frac{\beta_j}{\delta}z_j} \label{eq: PatMat dual function z} \\
        & - \delta \nneg \tau. \label{eq: PatMat dual function delta}
    \end{align}
  \end{subequations}

  Note that the resulting dual function is very similar to one~\eqref{eq: TopPushK dual function} for \TopPushK. In fact, the first three parts of~\eqref{eq: TopPushK dual function} and~\eqref{eq: PatMat dual function} are identical. Therefore, we only have to show how to minimize~\eqref{eq: PatMat dual function} with respect to~$\bm{z}.$ For that, we can use the conjugate function as in the case of minimization of~\eqref{eq: TopPushK dual function} with respect to~$\bm{y}.$  Then, for all $j = 1, \ldots, \; \nneg,$ we get 
  \begin{equation*}
    \delta \min_{\bm{z}} \; \Brac{l(\vartheta z_j) - \frac{\beta_j}{\delta\vartheta } \vartheta z_j} = - \delta l^{\star} \Brac{\frac{\beta_i}{\delta\vartheta }},
  \end{equation*}
  where the equality follows from Definition~\ref{def: conjugate} of a conjugate function. Plugging this back into~\eqref{eq: PatMat dual function z} yields the third part of the objective function~\eqref{eq: patmat family dual L}, which finishes the proof.
\end{proof}

\section{Coordinate Descent Algorithm}
\subsection{Family of \TopPushK Formulations}\label{sec: toppushk family coordinate proofs}
\subsubsection{Hinge Loss}

\topruleaa*
\begin{proof}[Proof of Proposition~\ref{prop: toppushk family hinge update a,a} on page~\pageref{prop: toppushk family hinge update a,a}]
  Constraint~\eqref{eq: Top dual hinge c1} is always satisfied from the definition of the update rule~\eqref{eq: update rule a,a}, and constraint~\eqref{eq: Top dual hinge c3} is always satisfied since no~$\beta_j$ was updated and the sum of all~$\alpha_i$ did not change. Constraint~\eqref{eq: Top dual hinge c2} reads
  \begin{align*}
    0 \leq \alphak + \Delta \leq C
    & \quad \implies \quad
    - \alphak \leq \Delta \leq C - \alphak, \\
    0 \leq \alphal - \Delta \leq C
    & \quad \implies \quad
    \alphal - C \leq \Delta \leq \alphal,
  \end{align*}
  which gives the lower and upper bound of~$\Delta.$
  
  Using the update rule~\eqref{eq: update rule a,a}, objective function~\eqref{eq: Top dual hinge L} can be rewritten as a quadratic function with respect to~$\Delta$
  \begin{equation*}
    - \frac{1}{2} \Brac[s]{\K_{kk} + \K_{ll} - \K_{kl} - \K_{lk}} \Delta^2
    - \Brac[s]{s_k - s_l} \Delta
    - c(\bm{\alpha}, \bm{\beta}).
  \end{equation*}
  Finally, the optimal solution~$\Delta^{\star}$ is given by~\eqref{eq: Delta optimal}.
\end{proof}

\topruleab*
\begin{proof}[Proof of Proposition~\ref{prop: toppushk family hinge update a,b} on page~\pageref{prop: toppushk family hinge update a,b}]
  Constraint~\eqref{eq: Top dual hinge c1} is always satisfied from the definition of the update rule~\eqref{eq: update rule a,b}. Constraint~\eqref{eq: Top dual hinge c2} reads~$- \alphak \leq \Delta \leq C - \alphak.$ Using the definition of~$\beta_{\max},$ constraint~\eqref{eq: Top dual hinge c3} for any~$K \geq 2$ reads
  \begin{align*}
    0 \leq \beta_{\max} \leq \frac{1}{K} \sum_{i = 1}^{\npos} \alpha_i + \frac{\Delta}{K} 
    & \quad \implies \quad
    K\beta_{\max} - \sum_{i = 1}^{\npos} \alpha_i \leq \Delta, \\
    0 \leq \betal + \Delta \leq \frac{1}{K} \sum_{i = 1}^{\npos} \alpha_i + \frac{\Delta}{K}
    & \quad \implies \quad
    -\betal \leq \Delta \quad \land \quad \Delta \leq \frac{1}{K-1}\Brac{\sum_{i = 1}^{\npos} \alpha_i - K \betal}.
  \end{align*}
  The combination of these bounds yields the lower bound~$\Delta_{lb}$ and upper bound~$\Delta_{ub}.$ If~$K = 1,$ the upper bound in~\eqref{eq: Top dual hinge c3} is always satisfied due to~\eqref{eq: Top dual hinge c1} and the lower and upper bound of~$\Delta$ can be simplified.
  
  Using the update rule~\eqref{eq: update rule a,b}, objective function~\eqref{eq: Top dual hinge L} can be rewritten as a quadratic function with respect to~$\Delta$
  \begin{equation*}
    - \frac{1}{2} \Brac[s]{\K_{kk} + \K_{ll} + \K_{kl} + \K_{lk}} \Delta^2
    - \Brac[s]{s_k + s_l - 1} \Delta
    - c(\bm{\alpha}, \bm{\beta}).
  \end{equation*}
  Finally, the optimal solution~$\Delta^{\star}$ is given by~\eqref{eq: Delta optimal}.
\end{proof}

\toprulebb*
\begin{proof}[Proof of Proposition~\ref{prop: toppushk family hinge update b,b} on page~\pageref{prop: toppushk family hinge update b,b}]
  Constraint~\eqref{eq: Top dual hinge c1} is always satisfied from the definition of the update rule~\eqref{eq: update rule b,b}, and constraint~\eqref{eq: Top dual hinge c2} is satisfied since no~$\alpha_i$ is updated. Constraint~\eqref{eq: Top dual hinge c3} for any~$K \geq 2$ reads
  \begin{align*}
    0 \leq \betak + \Delta \leq \frac{1}{K} \sum_{i = 1}^{\npos} \alpha_i 
    & \quad \implies \quad
    -\betak \leq \Delta \leq \frac{1}{K} \sum_{i = 1}^{\npos} \alpha_i - \betak, \\
    0 \leq \betal - \Delta \leq \frac{1}{K} \sum_{i = 1}^{\npos} \alpha_i
    & \quad \implies \quad
    \betal - \frac{1}{K} \sum_{i = 1}^{\npos} \alpha_i \leq \Delta \leq \betal,
  \end{align*}
  which gives the lower and upper bound of~$\Delta.$ If~$K = 1,$ the upper bound in~\eqref{eq: Top dual hinge c3} is always satisfied due to~\eqref{eq: Top dual hinge c1} and the lower and upper bound of~$\Delta$ can be simplified.
  
  Using the update rule~\eqref{eq: update rule b,b}, objective function~\eqref{eq: Top dual hinge L} can be rewritten as a quadratic function with respect to~$\Delta$
  \begin{equation*}
    - \frac{1}{2} \Brac[s]{\K_{kk} + \K_{ll} - \K_{kl} - \K_{lk}} \Delta^2
    - \Brac[s]{s_k - s_l} \Delta
    - c(\bm{\alpha}, \bm{\beta}).
  \end{equation*}
  Finally, the optimal solution~$\Delta^{\star}$ is given by~\eqref{eq: Delta optimal}.
\end{proof}

\subsubsection{Quadratic Hinge Loss}

The second considered surrogate function is the quadratic hinge loss from Notation~\ref{not: surrogates}. Plugging the conjugate~\eqref{eq: conjugate hinge} of the quadratic hinge loss into the dual formulation~\eqref{eq: toppushk family dual} yields
\begin{maxi!}{\bm{\alpha}, \bm{\beta}}{
  - \frac{1}{2} \vecab^\top \K \vecab
  + \sum_{i = 1}^{\npos} \alpha_i
  - \frac{1}{4C} \sum_{i = 1}^{\npos} \alpha_i^2
  }{\label{eq: Top dual quadratic}}{\label{eq: Top dual quadratic L}}
  \addConstraint{\sum_{i = 1}^{\npos} \alpha_i}{= \sum_{j = 1}^{\ntil} \beta_j
  \label{eq: Top dual quadratic c1}}
  \addConstraint{0 \leq \alpha_i}{,}{i = 1, 2, \ldots, \npos
  \label{eq: Top dual quadratic c2}}
  \addConstraint{0 \leq \beta_j}{\leq \frac{1}{K} \sum_{i = 1}^{\npos} \alpha_i, \quad}{j = 1, 2, \ldots, \ntil,
  \label{eq: Top dual quadratic c3}}
\end{maxi!}
Similarly to the previous case, the form of~$\K$ and~$\ntil$ depends on the used formulation and the upper bound in~\eqref{eq: Top dual quadratic c3} can be omitted for~$K = 1.$

\begin{proposition}[Update rule~\eqref{eq: update rule a,a} for problem~\eqref{eq: Top dual quadratic}]\label{prop: toppushk family quadratic update a,a}
  Consider problem~\eqref{eq: Top dual quadratic}, update rule~\eqref{eq: update rule a,a}, indeices~$1 \leq k \leq \npos$ and~$1 \leq l \leq \npos$ and Notation~\ref{not: dual update rules}. Then the optimal solution~$\Delta^{\star}$ is given by~\eqref{eq: Delta optimal} where
  \begin{align*}
    \Delta_{lb} & = -\alphak, &
    \Delta_{ub} & = \alphal, &
    \gamma & = -\frac{s_k - s_l + \frac{1}{2C}(\alphak - \alphal)}{\K_{kk} + \K_{ll} - \K_{kl} - \K_{lk} + \frac{1}{C}}.
  \end{align*}
\end{proposition}

\begin{proof}
  Constraint~\eqref{eq: Top dual quadratic c1} is always satisfied from the definition of the update rule~\eqref{eq: update rule a,a}. Constraint~\eqref{eq: Top dual quadratic c3} is also always satisfied since no~$\beta_j$ was updated and the sum of all~$\alpha_i$ did not change. Constraint~\eqref{eq: Top dual quadratic c2} reads
  \begin{align*}
    0 \leq \alphak + \Delta
    & \quad \implies \quad
    - \alphak \leq \Delta, \\
    0 \leq \alphal - \Delta
    & \quad \implies \quad
    \Delta \leq \alphal,
  \end{align*}
  which gives the lower and upper bound of~$\Delta.$
  
  Using the update rule~\eqref{eq: update rule a,a}, objective function~\eqref{eq: Top dual quadratic L} can be rewritten as a quadratic function with respect to~$\Delta$
  \begin{equation*}
    - \frac{1}{2} \Brac[s]{\K_{kk} + \K_{ll} - \K_{kl} - \K_{lk} + \frac{1}{C}} \Delta^2
    - \Brac[s]{s_k - s_l + \frac{1}{2C}(\alphak - \alphal)} \Delta
    - c(\bm{\alpha}, \bm{\beta}).
  \end{equation*}
  Finally, the optimal solution~$\Delta^{\star}$ is given by~\eqref{eq: Delta optimal}.
\end{proof}

\begin{proposition}[Update rule~\eqref{eq: update rule a,b} for problem~\eqref{eq: Top dual quadratic}]\label{prop: toppushk family quadratic update a,b}
  Consider problem~\eqref{eq: Top dual quadratic}, update rule~\eqref{eq: update rule a,b}, indeices~$1 \leq k \leq \npos$ and~$\npos + 1 \leq l \leq \ntil$  and Notation~\ref{not: dual update rules}. Let us define
  \begin{equation*}
    \beta_{\max} = \max_{j \in \{1, 2, \ldots, \ntil \} \setminus \{\hat{l}\}} \beta_j.
  \end{equation*}
  Then the optimal solution~$\Delta^{\star}$ is given by~\eqref{eq: Delta optimal} where
  \begin{align*}
    \Delta_{lb} & = 
      \begin{cases*}
        \max \Brac[c]{- \alphak,\;  -\betal} & K = 1, \\
        \max \Brac[c]{- \alphak,\;  -\betal, \; K\beta_{\max} - \sum_{i = 1}^{\npos} \alpha_i} & \textrm{otherwise},
      \end{cases*} \\
    \Delta_{ub} & = 
      \begin{cases*}
        + \infty & K = 1, \\
        \frac{1}{K-1}\Brac{\sum_{i = 1}^{\npos} \alpha_i - K \betal} & \textrm{otherwise},
      \end{cases*} \\
    \gamma & = -\frac{s_k + s_l - 1 + \frac{1}{2C} \alphak}{\K_{kk} + \K_{ll} + \K_{kl} + \K_{lk} + \frac{1}{2C}}.
  \end{align*}
\end{proposition}

\begin{proof}
  Constraint~\eqref{eq: Top dual quadratic c1} is always satisfied from the definition of the update rule~\eqref{eq: update rule a,b}. Constraint~\eqref{eq: Top dual quadratic c2} reads~$- \alphak \leq \Delta.$ Using the definition of~$\beta_{\max},$ constraint~\eqref{eq: Top dual quadratic c3} for any~$K \geq 2$ reads
  \begin{align*}
    0 \leq \beta_{\max} \leq \frac{1}{K} \sum_{i = 1}^{\npos} \alpha_i + \frac{\Delta}{K} 
    & \quad \implies \quad
    K\beta_{\max} - \sum_{i = 1}^{\npos} \alpha_i \leq \Delta, \\
    0 \leq \betal + \Delta \leq \frac{1}{K} \sum_{i = 1}^{\npos} \alpha_i + \frac{\Delta}{K}
    & \quad \implies \quad
    -\betal \leq \Delta \quad \land \quad \Delta \leq \frac{1}{K-1}\Brac{\sum_{i = 1}^{\npos} \alpha_i - K \betal}.
  \end{align*}
  The combination of these bounds yields the lower bound~$\Delta_{lb}$ and upper bound~$\Delta_{ub}.$ If~$K = 1,$ the upper bound in~\eqref{eq: Top dual quadratic c3} is always satisfied due to~\eqref{eq: Top dual quadratic c1} and the lower and upper bound of~$\Delta$ can be simplified.
  
  Using the update rule~\eqref{eq: update rule a,b}, objective function~\eqref{eq: Top dual quadratic L} can be rewritten as a quadratic function with respect to~$\Delta$
  \begin{equation*}
    - \frac{1}{2} \Brac[s]{\K_{kk} + \K_{ll} + \K_{kl} + \K_{lk} + \frac{1}{2C}} \Delta^2
    - \Brac[s]{s_k + s_l - 1 + \frac{1}{2C} \alphak} \Delta
    - c(\bm{\alpha}, \bm{\beta}).
  \end{equation*}
  Finally, the optimal solution~$\Delta^{\star}$ is given by~\eqref{eq: Delta optimal}.
\end{proof}

\begin{proposition}[Update rule~\eqref{eq: update rule b,b} for problem~\eqref{eq: Top dual quadratic}]\label{prop: toppushk family quadratic update b,b}
  Consider problem~\eqref{eq: Top dual quadratic}, update rule~\eqref{eq: update rule b,b}, indices~$\npos + 1 \leq k \leq \ntil$ and~$\npos + 1 \leq l \leq \ntil$  and Notation~\ref{not: dual update rules}. Then the optimal solution~$\Delta^{\star}$ is given by~\eqref{eq: Delta optimal} where
  \begin{align*}
    \Delta_{lb} & = 
      \begin{cases*}
        -\betak & K = 1, \\
        \max \Brac[c]{- \betak,\; \betal - \frac{1}{K} \sum_{i = 1}^{\npos} \alpha_i} & \textrm{otherwise},
      \end{cases*} \\
    \Delta_{ub} & = 
      \begin{cases*}
        \betal & K = 1, \\
        \min \Brac[c]{\betal,\; \frac{1}{K} \sum_{i = 1}^{\npos} \alpha_i - \betak} & \textrm{otherwise},
      \end{cases*} \\
    \gamma & = -\frac{s_k - s_l}{\K_{kk} + \K_{ll} - \K_{kl} - \K_{lk}}.
  \end{align*}
\end{proposition}

\begin{proof}
  Constraint~\eqref{eq: Top dual quadratic c1} is always satisfied from the definition of the update rule~\eqref{eq: update rule b,b}. Constraint~\eqref{eq: Top dual quadratic c2} is also always satisfied since no~$\alpha_i$ is updated. Constraint~\eqref{eq: Top dual quadratic c3} for any~$K \geq 2$ reads
  \begin{align*}
    0 \leq \betak + \Delta \leq \frac{1}{K} \sum_{i = 1}^{\npos} \alpha_i 
    & \quad \implies \quad
    -\betak \leq \Delta \leq \frac{1}{K} \sum_{i = 1}^{\npos} \alpha_i - \betak, \\
    0 \leq \betal - \Delta \leq \frac{1}{K} \sum_{i = 1}^{\npos} \alpha_i
    & \quad \implies \quad
    \betal - \frac{1}{K} \sum_{i = 1}^{\npos} \alpha_i \leq \Delta \leq \betal,
  \end{align*}
  which gives the lower and upper bound of~$\Delta.$ If~$K = 1,$ the upper bound in~\eqref{eq: Top dual quadratic c3} is always satisfied due to~\eqref{eq: Top dual quadratic c1} and the lower and upper bound of~$\Delta$ can be simplified.
  
  Using the update rule~\eqref{eq: update rule b,b}, objective function~\eqref{eq: Top dual quadratic L} can be rewritten as a quadratic function with respect to~$\Delta$
  \begin{equation*}
    - \frac{1}{2} \Brac[s]{\K_{kk} + \K_{ll} - \K_{kl} - \K_{lk}} \Delta^2
    - \Brac[s]{s_k - s_l} \Delta
    - c(\bm{\alpha}, \bm{\beta}).
  \end{equation*}
  Finally, the optimal solution~$\Delta^{\star}$ is given by~\eqref{eq: Delta optimal}.
\end{proof}

\subsubsection{Initialization}

\topinit*
\begin{proof}[Proof of Theorem~\ref{thm: toppushk family initialization} on page~\pageref{thm: toppushk family initialization}]
  The Lagrangian of~\eqref{eq: toppushk family initialization} reads
  \begin{align*}
    \mathcal{L}(\bm{\alpha}, \bm{\beta}; \lambda, \bm{p}, \bm{q}, \bm{u}, \bm{v})
     = \frac{1}{2} \norm{\bm{\alpha} - \bm{\alpha}^0}^2
     + \frac{1}{2} \norm{\bm{\beta} - \bm{\beta}^0}^2
     + \lambda \Brac{\sum_{i = 1}^{\npos} \alpha_i - \sum_{j = 1}^{\ntil} \beta_j} \\
     - \sum_{i = 1}^{\npos} p_i \alpha_i
     + \sum_{i = 1}^{\npos} q_i (\alpha_i - C_1)
     - \sum_{j = 1}^{\ntil} u_j \beta_j
     + \sum_{j = 1}^{\ntil} v_j \Brac{\beta_j - \frac{1}{K} \sum_{i = 1}^{\npos} \alpha_i}.
  \end{align*}
  The KKT conditions then amount to
  \begin{subequations}\label{eq:problem3_KKT}
  \begin{align}
    \frac{\partial \mathcal{L}}{\partial \alpha_i}
      & = \alpha_i - \alpha_i^0 + \lambda - p_i + q_i - \frac{1}{K} \sum_{j=1}^{\ntil} v_j = 0,
      && i = 1, 2, \ldots, \npos, \label{eq:problem3_KKT_opt1}\\
    \frac{\partial \mathcal{L}(\cdot)}{\partial \beta_j}
      & = \beta_j- \beta_j^0 - \lambda - u_j + v_j = 0,
      && j = 1, 2, \ldots, \ntil, \label{eq:problem3_KKT_opt2}
  \end{align}
  the primal feasibility conditions~\eqref{eq: toppushk family initialization}, the dual feasibility conditions $\lambda \in \R$, $p_i \ge 0$, $q_i \ge 0$, $u_j \ge 0$, $v_j \ge 0$ and finally the complementarity conditions
  \begin{align}
    p_i \alpha_i & = 0,
      && i = 1, 2, \ldots, \npos, \label{eq:problem3_KKT_comp1} \\
    q_i \Brac{\alpha_i - C_1} & = 0,
      && i = 1, 2, \ldots, \npos, \label{eq:problem3_KKT_comp2} \\
    u_j \beta_j & = 0,
      && j = 1, 2, \ldots, \ntil, \label{eq:problem3_KKT_comp3} \\
    v_j \Brac{\beta_j -\frac{1}{K} \sum_{i=1}^{\npos} \alpha_i} & =0,
      && j = 1, 2, \ldots, \ntil. \label{eq:problem3_KKT_comp4}
  \end{align}
  \end{subequations}
  
  \paragraph*{Case 1:} The first case concerns when the optimal solution satisfies~$\sum_i  \alpha_i = 0$. From the primal feasibility conditions, we immediately get~$\alpha_i = 0$ for all~$i$ and~$\beta_j = 0$ for all~$j$. Then~\eqref{eq:problem3_KKT_comp2} implies~$q_i = 0$ for all~$i$ and all complementarity conditions are satisfied. Moreover, optimality condition~\eqref{eq:problem3_KKT_opt1} implies
  \begin{equation*}
    \lambda = \alpha_i^0 + p_i + \frac{1}{K} \sum_{j = 1}^{\ntil} v_j.
  \end{equation*}
  Since the only condition on $p_i$ is the non-negativity, this implies
  \begin{equation*}
    \lambda \ge \max_{i=1, \ldots, \npos} \alpha_i^0 + \frac{1}{K} \sum_{j = 1}^{\ntil} v_j.
  \end{equation*}

  Similarly, from optimality condition~\eqref{eq:problem3_KKT_opt2} we deduce
  \begin{equation*}
    v_j
      = \beta_j^0 + \lambda + u_j
      \ge \beta_j^0 + \lambda
      \ge \beta_j^0 + \max_{i=1,\dots,\npos} \alpha_i^0 + \frac{1}{K} \sum_{i = 1}^{\ntil} v_i.
  \end{equation*}
  Since we need to fulfill $v_j \ge 0$, this amounts to
  \begin{equation*}
    v_j 
      \ge \clip[u]{0}{+\infty}{\beta_j^0 + \max_{i=1,\dots,\npos} \alpha_i^0 + \frac{1}{K} \sum_{i = 1}^{\ntil} v_i}.
  \end{equation*}
  Summing this with respect to $j$ and using the substitution $\bar{v} = \frac{1}{K}\sum_i v_i$ results in
  \begin{equation}\label{eq:problem3_proof0}
    K\bar{v} - \sum_{j=1}^{\ntil} \clip[u]{0}{+\infty}{\beta_j^0 + \max_{i=1,\dots,\npos} \alpha_i^0 + \bar{v}} \geq 0.
  \end{equation}
  Denote by~$\beta_{[j]}^0$ the sorted version of~$\beta_j^0$. Then the function on the left-hand side of~\eqref{eq:problem3_proof0} as a function of~$\bar{v}$ is increasing on~$\left(-\infty, \; -\beta_{[\npos - K + 1]}^0 - \max_{i} \alpha_i^0 \right]$ and non-increasing otherwise. Thus,~\eqref{eq:problem3_proof0} can be satisfied if and only if its function value at~$- \beta_{[\npos - K + 1]}^0 - \max_{i} \alpha_i^0$ is non-negative
  \begin{multline*}
    K\Brac{- \beta_{[\npos - K + 1]}^0 - \max_{i=1,\dots,\npos} \alpha_i^0} - \sum_{j=1}^{\ntil} \clip[u]{0}{+\infty}{\beta_j^0 + \max_{i=1,\dots,\npos} \alpha_i^0 - \beta_{[\npos - K + 1]}^0 - \max_{i=1,\dots,\npos} \alpha_i^0} \\
    = K\Brac{- \beta_{[\npos - K + 1]}^0 - \max_{i=1,\dots,\npos} \alpha_i^0} - \sum_{j=1}^{K} \Brac{\beta_{[\npos - K + j]}^0 - \beta_{[\npos - K + 1]}^0}
    =  - \sum_{j=1}^{K} \Brac{\beta_{[\npos - K + j]}^0 + \max_{i=1,\dots,\npos} \alpha_i^0} \geq 0,
  \end{multline*}
  which is precisely condition~\eqref{eq:problem3_cond}.
  
  \paragraph*{Case 2:} If~\eqref{eq:problem3_cond} holds true, then from the discussion above we obtain that the optimal solution satisfies~$\sum_i \alpha_i > 0$. For simplicity, we define
  \begin{align*}
    \bar{\alpha} & = \frac{1}{K} \sum_{i=1}^{\npos} \alpha_i, &
    \bar{\beta} & = \frac{1}{K} \sum_{j=1}^{\ntil} \beta_j, &
    \bar{v} & = \frac{1}{K} \sum_{j=1}^{\ntil} v_j.
  \end{align*}
  For any fixed~$i$, the standard trick is to combine the optimality condition~\eqref{eq:problem3_KKT_opt1} with the primal feasibility condition~$0 \le \alpha_i \le C_1$, the dual feasibility conditions $p_i \ge 0$, $q_i \ge 0$ and the complementarity conditions~(\ref{eq:problem3_KKT_comp1}, \ref{eq:problem3_KKT_comp2}) to obtain
  \begin{equation}\label{eq:problem3_alpha}
    \alpha_i = \clip{0}{C_1}{\alpha_i^0 - \lambda + \bar{v}}.
  \end{equation}
  
  Similarly for any fixed~$j$, we combine the optimality condition~\eqref{eq:problem3_KKT_opt2} with the primal feasibility condition~$0 \le \beta_j \le \bar{\alpha}$, the dual feasibility conditions $u_j \ge 0$, $v_j \ge 0$ and the complementarity conditions~(\ref{eq:problem3_KKT_comp3}, \ref{eq:problem3_KKT_comp4}) to obtain
  \begin{align}
    \beta_j
      & = \clip{0}{\bar{\alpha}}{\beta_j^0 + \lambda}, \label{eq:problem3_beta} \\
    v_j
      & = \clip[u]{0}{+\infty}{\beta_j^0 + \lambda - \bar{\alpha}}. \label{eq:problem3_rho}
  \end{align}
  Summing equations~\eqref{eq:problem3_alpha},~\eqref{eq:problem3_beta} and~\eqref{eq:problem3_rho} respectively with respect to~$i$ and~$j$ results in
  \begin{subequations}
    \begin{align}
      K \bar{\alpha}
        & = \sum_{i=1}^{\npos}\clip{0}{C_1}{\alpha_i^0 - \lambda + \bar{v}},
        \label{eq:problem3_proof1} \\
      K \bar{\beta}
        &= \sum_{j=1}^{\ntil} \clip{0}{\bar{\alpha}}{\beta_j^0 + \lambda},
        \label{eq:problem3_proof2} \\
      K \bar{v}
        & = \sum_{j=1}^{\ntil} \clip[u]{0}{+\infty}{\beta_j^0 + \lambda - \bar{\alpha}}.
        \label{eq:problem3_proof3}
    \end{align}
  \end{subequations}
  We denote $\mu = \bar{\alpha}$. Then~\eqref{eq: toppushk family init alg 1} results by plugging~\eqref{eq:problem3_proof3} into~\eqref{eq:problem3_proof1} while~\eqref{eq: toppushk family init alg 2} follows from~\eqref{eq:problem3_proof2} and $\sum_i \alpha_i = \sum_j \beta_j$. 
\end{proof}

\topinith*
\begin{proof}[Proof of Lemma~\ref{lemma: toppushk family h} on page~\pageref{lemma: toppushk family h}]
  Recall that based on~\eqref{eq: toppushk family init alg 2} we defined
  \begin{equation*}
    g(\lambda; \mu) := \sum_{j=1}^{\ntil} \clip{0}{\mu}{\beta_j^0 + \lambda} - K \mu,
  \end{equation*}
  and solutions of~$g(\lambda; \mu) = 0$ for a fixed~$\mu$ are denoted by~$\lambda(\mu)$.
  
  Let us first consider the case, when the solution to~$g(\lambda) = 0$ is not unique. Since function~$g(\cdot; \, \mu)$ is non-decreasing and~$K$ is an integer, it can happen only if the solution~$\lambda(\mu)$ satisfies
  \begin{equation*}
    \beta_{[j]}^0 + \lambda(\mu) \;
    \begin{cases}
      \ge \mu & \text{for } j = \ntil - K + 1, \dots , \ntil, \\
      \le 0 & \text{otherwise.}
    \end{cases}
  \end{equation*}
  Here, we again denote~$\bm{\beta}_{[\cdot]}^0$ to be the sorted version of~$\bm{\beta}_j^0$. Then~$h$ defined in~\eqref{eq: toppushk family h} equals to
  \begin{align*}
    h(\mu)
      & = \sum_{i=1}^{\npos} \clip{0}{C_1}{\alpha_i^0 - \lambda(\mu) + \frac{1}{K} \sum_{j=\ntil - K + 1}^{\ntil} \Brac{\beta_j^0+\lambda(\mu) - \mu}} - K \mu \\
      & = \sum_{i=1}^{\npos} \clip{0}{C_1}{\alpha_i^0 - \mu + \frac{1}{K} \sum_{j=\ntil - K + 1}^{\ntil} \beta_j^0} - K \mu.
  \end{align*}
  This implies the first statement of the lemma that~$h$ is independent of the choice of~$\lambda(\mu)$.
  
  In the previous paragraph, we prove, that~$h$ gives the same value for every choice of~$\lambda(\mu).$ Now we need to show that~$h$ is a decreasing function for the arbitrary choice of~$\lambda(\mu).$ Fix any~$\mu_2 > \mu_1 > 0$. From~\eqref{eq: toppushk family init alg 2} we have
  \begin{align}
   \sum_{j=1}^{\ntil} \clip{0}{\mu_1}{\beta_j^0 + \lambda(\mu_1)} - K \mu_1 & = 0,
    \label{eq:system2_proof1} \\
  \sum_{j=1}^{\ntil} \clip{0}{\mu_2}{\beta_j^0 + \lambda(\mu_2)} - K \mu_2 & = 0.
    \label{eq:system2_proof2}
  \end{align}
  Equation~\eqref{eq:system2_proof1} implies that at most~$K$ values of~$\beta_j^0 + \lambda(\mu_1)$ are greater or equal than~$\mu_1$. If we increase the upper bound in the projection, at most~$K$ values can increase, which results in
  \begin{equation}\label{eq:system2_proof3}
    \sum_{j=1}^{\ntil} \clip{0}{\mu_2}{\beta_j^0 + \lambda(\mu_1)}
      \le \sum_{j=1}^{\ntil} \clip{0}{\mu_1}{\beta_j^0 + \lambda(\mu_1)} + K(\mu_2 - \mu_1)
      = K \mu_2,
  \end{equation}
  where the equality follows from~\eqref{eq:system2_proof1}. Comparing~\eqref{eq:system2_proof2} and~\eqref{eq:system2_proof3} yields~$\lambda(\mu_2) \ge \lambda(\mu_1)$.
  
  Now define
  \begin{equation*}
    J = \Set{j}{\beta_j^0 + \lambda(\mu_1) \ge 0}
  \end{equation*}
  and observe that due to~\eqref{eq:system2_proof1} we have~$\abs{J} \ge K$. Moreover, the definition of $J$ and~\eqref{eq:system2_proof1} yields
  \begin{equation}\label{eq:system2_proof4}
    \sum_{j \in J} \clip{0}{\mu_1}{\beta_j^0 + \lambda(\mu_1)} - K\mu_1
      = \sum_{j=1}^{\ntil} \clip{0}{\mu_1}{\beta_j^0 + \lambda(\mu_1)} - K\mu_1
      = 0.
  \end{equation}
  Then we have
  \begin{align*}
    \sum_{j=1}^{\ntil} \clip{0}{\mu_2}{\beta_j^0 + \lambda(\mu_1) + \mu_2 - \mu_1}
      & \ge \sum_{j \in J} \clip{0}{\mu_2}{\beta_j^0 + \lambda(\mu_1) + \mu_2 - \mu_1} \\
      &  = \sum_{j \in J} \clip{\mu_2 - \mu_1}{\mu_2}{\beta_j^0 + \lambda(\mu_1) + \mu_2 - \mu_1} \\
      & = \sum_{j \in J} \clip{0}{\mu_1}{\beta_j^0 + \lambda(\mu_1)} + \abs{J}(\mu_2 - \mu_1) \\
      &  = K\mu_1 + \abs{J}(\mu_2 - \mu_1)
      \ge K\mu_1 + K(\mu_2 - \mu_1)
      = K\mu_2,
  \end{align*}
  where the first equality follows from the definition of~$J$ and the second equality is a shift by a~$\mu_2- \mu_1.$ The third equality follows from~\eqref{eq:system2_proof4} and finally, the last inequality follows from~$\abs{J} \ge K$. The chain above together with~\eqref{eq:system2_proof2} implies~$\lambda(\mu_2) - \mu_2 \le \lambda(\mu_1)- \mu_1$. Combining this with~$\mu_2 > \mu_1$ and~$\lambda(\mu_2) \ge \lambda(\mu_1)$, this implies that~$h$ from~\eqref{eq: toppushk family h} is non-increasing which is precisely the lemma statement.
\end{proof}

\subsection{Family of \PatMat Formulations}\label{sec: Pat coordinate descent}

In this section, we derive a coordinate descent algorithm for solving dual formulation~\eqref{eq: patmat family dual} for the family of \PatMat formulations. We follow the same approach as for \TopPushK family in Section\ref{sec: Top coordinate descent}, i.e. we use update rules~\eqref{eq: update rules}. In this case, we must also consider the third primary variable~$\delta.$ Then the dual formulation~\eqref{eq: patmat family dual}  can be rewritten as a one-dimensional quadratic problem
\begin{maxi*}{\Delta}{
  -\frac{1}{2} a(\bm{\alpha}, \bm{\beta}, \delta) \Delta^2
  - b(\bm{\alpha}, \bm{\beta}, \delta) \Delta
  - c(\bm{\alpha}, \bm{\beta}, \delta)
  }{}{}
  \addConstraint{\Delta_{lb}(\bm{\alpha}, \bm{\beta}, \delta)}{\leq \Delta \leq \Delta_{ub}(\bm{\alpha}, \bm{\beta}, \delta)}
\end{maxi*}
where~$a,$~$b,$~$c,$~$\Delta_{lb},$~$\Delta_{ub}$ are constants with respect to~$\Delta.$ The form of the optimal solution is the same as for problem~\eqref{eq: toppushk family dual} and reads
\begin{equation*}
  \Delta^{\star} = \clip{\Delta_{lb}}{\Delta_{ub}}{\gamma}.
\end{equation*}
Since we assume one of the update rule~\eqref{eq: update rules}, the constrain~\eqref{eq: patmat family dual c1} is always satisfied after the update. The exact form of the update rules depends on the surrogate function. Moreover, the form of optimal~$\delta$ also depends on the surrogate function. The upcoming text follows the same order as in the previous section. Therefore, we introduce concrete forms of update rules for hinge and quadratic hinge loss function and then show how to find an initial feasible solution.

\subsubsection{Hinge Loss}

We again start with the hinge loss function from Notation~\ref{not: surrogates}. Plugging the conjugate~\eqref{eq: conjugate hinge} of the hinge loss into the dual formulation~\eqref{eq: patmat family dual} yields
\begin{maxi!}{\bm{\alpha}, \bm{\beta}, \delta}{
  - \frac{1}{2} \vecab^\top \K \vecab
  + \sum_{i = 1}^{\npos} \alpha_i
  + \frac{1}{\vartheta} \sum_{j = 1}^{\ntil} \beta_j 
  - \delta \ntil \tau
  }{\label{eq: Pat dual hinge}}{\label{eq: Pat dual hinge L}}
  \addConstraint{\sum_{i = 1}^{\npos} \alpha_i}{= \sum_{j = 1}^{\ntil} \beta_j \label{eq: Pat dual hinge c1}}
  \addConstraint{0 \leq \alpha_i}{\leq C,}{i = 1, 2, \ldots, \npos \label{eq: Pat dual hinge c2}}
  \addConstraint{0 \leq \beta_j}{\leq \delta \vartheta, \quad}{j = 1, 2, \ldots, \ntil \label{eq: Pat dual hinge c3}}
  \addConstraint{\delta }{\geq 0. \label{eq: Pat dual hinge c4}}
\end{maxi!}
Since we know the form of the optimal solution~\eqref{eq: Delta optimal}, we only need to show how to compute~$\Delta_{lb},$~$\Delta_{ub}$ and~$\gamma$ for all update rules~\eqref{eq: update rules}. However, in this case, constants~$\Delta_{lb},$~$\Delta_{ub}$ and~$\gamma$ also depend on the third dual variable~$\delta$. We do not perform a joint maximization in~$(\alphak, \; \betal, \; \delta)$ but perform a maximization with respect to~$(\alphak, \; \betal)$, update these two values and then optimize the objective with respect to~$\delta$. Then for fixed feasible solution~$\bm{\alpha}$ and~$\bm{\beta},$ maximizing objective function~\eqref{eq: Pat dual hinge L} with respect to~$\delta$ yields
\begin{maxi*}{\delta}{
  - \ntil \tau \delta
  }{}{}
  \addConstraint{0 \leq \beta_j}{\leq \delta \vartheta, \quad}{j = 1, 2, \ldots, \ntil}
  \addConstraint{\delta \geq 0.}
\end{maxi*}
Since~$\ntil \tau \geq 0,$ we have to find the smallest possible~$\delta$ that satisfies constraints above. Such~$\delta$ is in the following form
\begin{equation}\label{eq: Pat dual hinge optimal delta}
  \delta^* = \frac{1}{\vartheta} \max_{j \in \{1, 2, \ldots, \ntil \}} \beta_j.
\end{equation}

The following three propositions provide closed-form formulae for all three update rules.

\begin{proposition}[Update rule~\eqref{eq: update rule a,a} for problem~\eqref{eq: Pat dual hinge}]\label{thm: patmat family hinge update a,a}
  Consider problem~\eqref{eq: Pat dual hinge}, update rule~\eqref{eq: update rule a,a}, indices~$1 \leq k \leq \npos$ and~$1 \leq l \leq \npos$  and Notation~\ref{not: dual update rules}. Then the optimal solution~$\Delta^{\star}$ is given by~\eqref{eq: Delta optimal} where
  \begin{align*}
    \Delta_{lb} & = \min\{- \alphak,\; \alphal - C\}, &
    \Delta_{ub} & = \max\{C - \alphak,\; \alphal\}, \\
    \gamma & = -\frac{s_k - s_l}{\K_{kk} + \K_{ll} - \K_{kl} - \K_{lk}}, &
    \delta^{\star} & = \delta.
  \end{align*}
\end{proposition}

\begin{proof}
  Constraint~\eqref{eq: Pat dual hinge c1} is always satisfied from the definition of the update rule~\eqref{eq: update rule a,a}. Constraint~\eqref{eq: Pat dual hinge c3} is also always satisfied since no~$\beta_j$ was updated and the sum of all~$\alpha_i$ did not change. Constraint~\eqref{eq: Pat dual hinge c2} reads
  \begin{align*}
    0 \leq \alphak + \Delta \leq C
    & \quad \implies \quad
    - \alphak \leq \Delta \leq C - \alphak \\
    0 \leq \alphal - \Delta \leq C
    & \quad \implies \quad
    \alphal - C \leq \Delta \leq \alphal
  \end{align*}
  which gives the lower and upper bound of~$\Delta.$
  
  Using the update rule~\eqref{eq: update rule a,a}, objective function~\eqref{eq: Pat dual hinge L} can be rewritten as a quadratic function with respect to~$\Delta$
  \begin{equation*}
    - \frac{1}{2} \Brac[s]{\K_{kk} + \K_{ll} - \K_{kl} - \K_{lk}} \Delta^2
    - \Brac[s]{s_k - s_l} \Delta
    - c(\bm{\alpha}, \bm{\beta}).
  \end{equation*}
  The optimal solution~$\Delta^{\star}$ is given by~\eqref{eq: Delta optimal}. Finally, since optimal~$\delta$ is given by~\eqref{eq: Pat dual hinge optimal delta} and no~$\beta_j$ was updated, the optimal~$\delta$ does not change.
\end{proof}

\begin{proposition}[Update rule~\eqref{eq: update rule a,b} for problem~\eqref{eq: Pat dual hinge}]\label{thm: patmat family hinge update a,b}
  Consider problem~\eqref{eq: Pat dual hinge}, update rule~\eqref{eq: update rule a,b}, indices~$1 \leq k \leq \npos$ and~$\npos + 1 \leq l \leq \ntil$  and Notation~\ref{not: dual update rules}. Let us define
  \begin{equation*}
    \beta_{\max} = \max_{j \in \{1, 2, \ldots, \ntil \} \setminus \{\hat{l}\}} \beta_j.
  \end{equation*}
  Then the bounds from~\eqref{eq: Delta optimal} are defined as~$\Delta_{lb} = \max\{- \alphak,\; -\betal \}$ and~$\Delta_{ub} = C - \alphak$ and there are two possible solutions
  \begin{enumerate}
    \item $\Delta^{\star}_1$ is feasible if~$\betal + \Delta^{\star}_1 \leq \beta_{\max}$ and is given by~\eqref{eq: Delta optimal} where
    \begin{align*}
      \gamma
        & = -\frac{s_k + s_l - 1 - \frac{1}{\vartheta}}{\K_{kk} + \K_{ll} + \K_{kl} + \K_{lk}}, &
      \delta^{*}_1
        & = \frac{\beta_{\max}}{\vartheta}.
    \end{align*}
    \item $\Delta^{\star}_2$ is feasible if~$\betal + \Delta^{\star}_2 \geq \beta_{\max}$ and is given by~\eqref{eq: Delta optimal} where
    \begin{align*}
      \gamma
        & = -\frac{s_k + s_l - 1 - \frac{1 - \ntil \tau}{\vartheta}}{\K_{kk} + \K_{ll} + \K_{kl} + \K_{lk}}, &
      \delta^{*}_2
        & = \frac{\betal + \Delta^{\star}_2}{\vartheta}.
    \end{align*}
  \end{enumerate}
  The optimal solution~$\Delta^{\star}$ is equal to one of them, which maximizes the original objective and is feasible.
\end{proposition}

\begin{proof}
  Constraint~\eqref{eq: Pat dual hinge c1} is always satisfied from the definition of the update rule~\eqref{eq: update rule a,b}. Constraint~\eqref{eq: Pat dual hinge c2} reads~$- \alphak \leq \Delta \leq C - \alphak.$ Using the definition of~$\beta_{\max},$ constraint~\eqref{eq: Pat dual hinge c3} reads~$\beta_{\max} \leq \delta \vartheta$ and~$0 \leq \betal + \Delta \leq \delta \vartheta.$ Since the optimal~$\delta$ is given by~\eqref{eq: Pat dual hinge optimal delta}, there are only two possible choices: $\delta^{\star}_1 = \frac{\beta_{\max}}{\vartheta}$ and~$\delta^{\star}_2 = \frac{\betal + \Delta}{\vartheta}.$ If~$\delta$ is feasible, all upper bounds in constraint~\eqref{eq: Pat dual hinge c3} hold. Therefore, we can simplify the constraints to~$- \betal \leq \Delta,$ which in combination with bounds for~$\alphak$ gives the lower and upper bound of~$\Delta.$ Now let us discuss how to select optimal~$\delta:$
  \begin{enumerate}
    \item Using~$\delta^{\star}_1$ and the update rule~\eqref{eq: update rule a,b}, objective function~\eqref{eq: Pat dual hinge L} can be rewritten as a quadratic function with respect to~$\Delta$ as
    \begin{equation*}
      - \frac{1}{2} \Brac[s]{\K_{kk} + \K_{ll} + \K_{kl} + \K_{lk}} \Delta^2
      - \Brac[s]{s_k + s_l - 1 - \frac{1}{\vartheta}} \Delta
      - c(\bm{\alpha}, \bm{\beta}).
    \end{equation*}
    The optimal solution~$\Delta^{\star}_1$ is given by~\eqref{eq: Delta optimal} and is feasible if~$\betal + \Delta^{\star}_1 \leq \beta_{\max}$.

    \item Using~$\delta^{\star}_2$ and the update rule~\eqref{eq: update rule a,b}, objective function~\eqref{eq: Pat dual hinge L} can be rewritten as a quadratic function with respect to~$\Delta$ as
    \begin{equation*}
      - \frac{1}{2} \Brac[s]{\K_{kk} + \K_{ll} + \K_{kl} + \K_{lk}} \Delta^2
      - \Brac[s]{s_k + s_l - 1 - \frac{1 - \ntil \tau}{\vartheta}} \Delta
      - c(\bm{\alpha}, \bm{\beta}).
    \end{equation*}
    The optimal solution~$\Delta^{\star}_2$ is given by~\eqref{eq: Delta optimal} and is feasible if~$\betal + \Delta^{\star}_2 \geq \beta_{\max}$.
  \end{enumerate}
  The optimal solution is the one, which maximizes the objective~\eqref{eq: Pat dual hinge L} and is feasible.
\end{proof}

\begin{proposition}[Update rule~\eqref{eq: update rule b,b} for problem~\eqref{eq: Pat dual hinge}]\label{thm: patmat family hinge update b,b}
  Consider problem~\eqref{eq: Pat dual hinge}, update rule~\eqref{eq: update rule b,b}, indices~$\npos + 1 \leq k \leq \ntil$ and~$\npos + 1 \leq l \leq \ntil$ and Notation~\ref{not: dual update rules}. Let us define
  \begin{equation*}
    \beta_{\max} = \max_{j \in \{1, 2, \ldots, \ntil \} \setminus \{\hat{k}, \hat{l}\}} \beta_j.
  \end{equation*}
  Then the bounds from~\eqref{eq: Delta optimal} are defined as~$\Delta_{lb} = - \betak$ and~$\Delta_{ub} = \betal$ and there are three possible solutions
  \begin{enumerate}
    \item $\Delta^{\star}_1$ is feasible if~$\beta_{\max} \geq \max\{\betak + \Delta^{\star}_1, \betal - \Delta^{\star}_1\}$ and is given by~\eqref{eq: Delta optimal} where
    \begin{align*}
      \gamma
        & = -\frac{s_k - s_l}{\K_{kk} + \K_{ll} - \K_{kl} - \K_{lk}}, &
      \delta^{*}_1
        & = \frac{\beta_{\max}}{\vartheta}.
    \end{align*}
    \item $\Delta^{\star}_2$ is feasible if~$\betak + \Delta^{\star}_2 \geq \max\{\beta_{\max} , \betal - \Delta^{\star}_2\}$ and is given by~\eqref{eq: Delta optimal} where
    \begin{align*}
      \gamma
        & = -\frac{s_k - s_l + \frac{\ntil \tau}{\vartheta}}{\K_{kk} + \K_{ll} - \K_{kl} - \K_{lk}}, &
      \delta^{*}_2
        & = \frac{\betak + \Delta^{\star}_2}{\vartheta}.
    \end{align*}
    \item $\Delta^{\star}_3$ is feasible if~$\betal - \Delta^{\star}_3 \geq \max\{\betak + \Delta^{\star}_3, \beta_{\max}\}$ and is given by~\eqref{eq: Delta optimal} where
    \begin{align*}
      \gamma
        & = -\frac{s_k - s_l - \frac{\ntil \tau}{\vartheta}}{\K_{kk} + \K_{ll} - \K_{kl} - \K_{lk}}, &
      \delta^{*}_3
        & = \frac{\betal - \Delta^{\star}_3}{\vartheta}.
    \end{align*}
  \end{enumerate}
  The optimal solution~$\Delta^{\star}$ is equal to one of them, which maximizes the original objective and is feasible.
\end{proposition}

\begin{proof}
  Constraint~\eqref{eq: Pat dual hinge c1} is always satisfied from the definition of the update rule~\eqref{eq: update rule b,b}. Constraint~\eqref{eq: Pat dual hinge c2} is also always satisfied since no~$\alpha_i$ is updated. Using the definition of~$\beta_{\max},$ constraint~\eqref{eq: Pat dual hinge c3} reads
  \begin{align*}
    \beta_{\max} & \leq \delta \vartheta, \\
    0 \leq \betak + \Delta & \leq \delta \vartheta, \\
    0 \leq \betal - \Delta & \leq \delta \vartheta.
  \end{align*}
  Since the optimal~$\delta$ is given by~\eqref{eq: Pat dual hinge optimal delta}, there are only two possible choices
  \begin{align}\label{eq: Pat dual hinge b, b proof delta}
    \delta^{\star}_1 & = \frac{\beta_{\max}}{\vartheta}, &
    \delta^{\star}_2 & = \frac{\betak + \Delta}{\vartheta}, &
    \delta^{\star}_3 & = \frac{\betal - \Delta}{\vartheta}.
  \end{align}
  If we use any of these choices which is feasible, all upper bounds in constraint~\eqref{eq: Pat dual hinge c3} hold, i.e. we can simplify the constraints to
  \begin{align*}
    0 \leq \betak + \Delta
    & \quad \implies \quad
    - \betak \leq \Delta, \\
    0 \leq \betal - \Delta
    & \quad \implies \quad
    \Delta \leq \betal,
  \end{align*}
  which gives the lower and upper bound of~$\Delta.$ Now let us discuss how to select optimal~$\delta:$
  \begin{enumerate}
    \item Using~$\delta^{\star}_1$ from~\eqref{eq: Pat dual hinge b, b proof delta} and the update rule~\eqref{eq: update rule b,b}, objective function~\eqref{eq: Pat dual hinge L} can be rewritten as a quadratic function with respect to~$\Delta$ as
    \begin{equation*}
      - \frac{1}{2} \Brac[s]{\K_{kk} + \K_{ll} - \K_{kl} - \K_{lk}} \Delta^2
      - \Brac[s]{s_k - s_l} \Delta
      - c(\bm{\alpha}, \bm{\beta}).
    \end{equation*}
    The optimal solution~$\Delta^{\star}_1$ is given by~\eqref{eq: Delta optimal} and is feasible if
    \begin{equation*}
      \beta_{\max} \geq \max\{\betak + \Delta^{\star}_1, \; \betal - \Delta^{\star}_1\}.
    \end{equation*}

    \item Using~$\delta^{\star}_2$ from~\eqref{eq: Pat dual hinge b, b proof delta} and the update rule~\eqref{eq: update rule b,b}, objective function~\eqref{eq: Pat dual hinge L} can be rewritten as a quadratic function with respect to~$\Delta$ as
    \begin{equation*}
      - \frac{1}{2} \Brac[s]{\K_{kk} + \K_{ll} - \K_{kl} - \K_{lk}} \Delta^2
      - \Brac[s]{s_k - s_l + \frac{\ntil \tau}{\vartheta}} \Delta
      - c(\bm{\alpha}, \bm{\beta}).
    \end{equation*}
    The optimal solution~$\Delta^{\star}_2$ is given by~\eqref{eq: Delta optimal} and is feasible if
    \begin{equation*}
      \betak + \Delta^{\star}_2 \geq \max\{\beta_{\max} , \betal - \Delta^{\star}_2\}.
    \end{equation*}

    \item Using~$\delta^{\star}_3$ from~\eqref{eq: Pat dual hinge b, b proof delta} and the update rule~\eqref{eq: update rule b,b}, objective function~\eqref{eq: Pat dual hinge L} can be rewritten as a quadratic function with respect to~$\Delta$ as
    \begin{equation*}
      - \frac{1}{2} \Brac[s]{\K_{kk} + \K_{ll} - \K_{kl} - \K_{lk}} \Delta^2
      - \Brac[s]{s_k - s_l - \frac{\ntil \tau}{\vartheta}} \Delta
      - c(\bm{\alpha}, \bm{\beta}).
    \end{equation*}
    The optimal solution~$\Delta^{\star}_3$ is given by~\eqref{eq: Delta optimal} and is feasible if
    \begin{equation*}
      \betal - \Delta^{\star}_3 \geq \max\{\beta_{\max}, \betak + \Delta^{\star}_3\}.
    \end{equation*}
  \end{enumerate}
  The optimal solution is the one, which maximizes the objective~\eqref{eq: Pat dual hinge L} and is feasible.
\end{proof}

\subsubsection{Quadratic Hinge Loss}

The second considered surrogate function is the quadratic hinge loss from Notation~\ref{not: surrogates}. Plugging the conjugate~\eqref{eq: conjugate quadratic hinge} of the quadratic hinge loss into the dual formulation~\eqref{eq: patmat family dual} yields
\begin{maxi!}{\bm{\alpha}, \bm{\beta}, \delta}{
  - \frac{1}{2} \vecab^\top \K \vecab
  + \sum_{i = 1}^{\npos} \alpha_i
  - \frac{1}{4C} \sum_{i = 1}^{\npos} \alpha_i^2
  }{\label{eq: Pat dual quadratic}}{\label{eq: Pat dual quadratic L1}}
  \breakObjective{
    + \frac{1}{\vartheta} \sum_{j = 1}^{\ntil} \beta_j 
    - \frac{1}{4 \delta \vartheta^2} \sum_{j = 1}^{\ntil} \beta_j^2
    - \delta \ntil \tau \label{eq: Pat dual quadratic L2}
  }
  \addConstraint{\sum_{i = 1}^{\npos} \alpha_i}{= \sum_{j = 1}^{\ntil} \beta_j
  \label{eq: Pat dual quadratic c1}}
  \addConstraint{\alpha_i}{\geq 0,}{i = 1, 2, \ldots, \npos
  \label{eq: Pat dual quadratic c2}}
  \addConstraint{\beta_j}{\geq 0,}{j = 1, 2, \ldots, \ntil
  \label{eq: Pat dual quadratic c3}}
  \addConstraint{\delta }{\geq 0,
  \label{eq: Pat dual quadratic c4}}
\end{maxi!}
Similar to the previous case, we perform maximization only with respect to~$(\alphak, \; \betal).$ Then for fixed feasible solution~$\bm{\alpha},$~$\bm{\beta},$ we need to maximize the objective function~(\ref{eq: Pat dual quadratic L1}-\ref{eq: Pat dual quadratic L2}) with respect to~$\delta$, which leads to the following problem
\begin{maxi*}{\delta}{
  - (\ntil \tau) \delta - \Brac{\frac{1}{4\vartheta^2} \sum_{j = 1}^{\ntil} \beta_j^2} \frac{1}{\delta}
  }{}{}
  \addConstraint{\delta \geq 0,}
\end{maxi*}
with the optimal solution that equals to
\begin{equation}\label{eq: Pat dual quadratic optimal delta}
  \delta^* = \sqrt{\frac{1}{4\vartheta^2 \ntil \tau} \sum_{j = 1}^{\ntil} \beta_j^2}.
\end{equation}

The following three propositions provide closed-form formulae for all three update rules.

\begin{proposition}[Update rule~\eqref{eq: update rule a,a} for problem~\eqref{eq: Pat dual quadratic}]\label{thm: patmat family quadratic update a,a}
  Consider problem~\eqref{eq: Pat dual quadratic}, update rule~\eqref{eq: update rule a,a}, indices~$1 \leq k \leq \npos$ and~$1 \leq l \leq \npos$  and Notation~\ref{not: dual update rules}. Then the optimal solution~$\Delta^{\star}$ is given by~\eqref{eq: Delta optimal} where
  \begin{align*}
    \Delta_{lb} & = -\alphak, \\
    \Delta_{ub} & = \alphal, \\
    \gamma & = -\frac{s_k - s_l + \frac{1}{2C}(\alphak - \alphal)}{\K_{kk} + \K_{ll} - \K_{kl} - \K_{lk} + \frac{1}{C}}, \\
    \delta^{\star}  & = \delta.
  \end{align*}
\end{proposition}

\begin{proof}
  Constraint~\eqref{eq: Pat dual quadratic c1} is always satisfied from the definition of the update rule~\eqref{eq: update rule a,a}. Constraint~\eqref{eq: Pat dual quadratic c3} is also always satisfied since no~$\beta_j$ was updated. Constraint~\eqref{eq: Pat dual quadratic c2} reads
  \begin{align*}
    0 \leq \alphak + \Delta
    & \quad \implies \quad
    - \alphak \leq \Delta, \\
    0 \leq \alphal - \Delta
    & \quad \implies \quad
    \Delta \leq \alphal,
  \end{align*}
  which gives the lower and upper bound of~$\Delta.$
  
  Using the update rule~\eqref{eq: update rule a,a}, objective function~(\ref{eq: Pat dual quadratic L1}-\ref{eq: Pat dual quadratic L2}) can be rewritten as a quadratic function with respect to~$\Delta$
  \begin{equation*}
    - \frac{1}{2} \Brac[s]{\K_{kk} + \K_{ll} - \K_{kl} - \K_{lk} + \frac{1}{C}} \Delta^2
    - \Brac[s]{s_k - s_l + \frac{1}{2C}(\alphak - \alphal)} \Delta
    - c(\bm{\alpha}, \bm{\beta}).
  \end{equation*}
  The optimal solution~$\Delta^{\star}$ is given by~\eqref{eq: Delta optimal}. Finally, since optimal~$\delta$ is given by~\eqref{eq: Pat dual quadratic optimal delta} and no~$\beta_j$ was updated, the optimal~$\delta$ does not change.
\end{proof}

\begin{proposition}[Update rule~\eqref{eq: update rule a,b} for problem~\eqref{eq: Pat dual quadratic}]\label{thm: patmat family quadratic update a,b}
  Consider problem~\eqref{eq: Pat dual quadratic}, update rule~\eqref{eq: update rule a,b}, indices~$1 \leq k \leq \npos$ and~$\npos + 1 \leq l \leq \ntil$ and Notation~\ref{not: dual update rules}. Then the optimal solution~$\Delta^{\star}$ is given by~\eqref{eq: Delta optimal} where
  \begin{align*}
    \Delta_{lb} & = \max\{- \alphak, - \betal\}, \\
    \Delta_{ub} & = +\infty, \\
    \gamma      & = -\frac{s_k + s_l  - 1 + \frac{\alphak}{2C} - \frac{1}{\vartheta} + \frac{\betal}{2 \delta \vartheta^2}}{\K_{kk} + \K_{ll} + \K_{kl} + \K_{lk} + \frac{1}{2C} + \frac{1}{2 \delta \vartheta^2}}, \\
    \delta^{\star}  & = \sqrt{\delta^2 + \frac{1}{4 \vartheta^2 \ntil \tau}({\Delta^{\star}}^2 + 2 \Delta^{\star} \betal)}.
  \end{align*}
\end{proposition}

\begin{proof}
  Constraint~\eqref{eq: Pat dual quadratic c1} is always satisfied from the definition of the update rule~\eqref{eq: update rule a,b}. Constraints~\eqref{eq: Pat dual quadratic c2} and~\eqref{eq: Pat dual quadratic c3} reads
  \begin{align*}
    0 \leq \alphak + \Delta
    & \quad \implies \quad
    - \alphak \leq \Delta, \\
    0 \leq \betal + \Delta
    & \quad \implies \quad
    - \betal \leq \Delta, \\
  \end{align*}
  which gives the lower bound of~$\Delta.$ In this case, $\Delta$ has no upper bound.
  
  Using the update rule~\eqref{eq: update rule a,b}, objective function~(\ref{eq: Pat dual quadratic L1}-\ref{eq: Pat dual quadratic L2}) can be rewritten as a quadratic function with respect to~$\Delta$
  \begin{align*}
    - \frac{1}{2} \Brac[s]{\K_{kk} + \K_{ll} + \K_{kl} + \K_{lk} + \frac{1}{2C} + \frac{1}{2 \delta \vartheta^2}} & \Delta^2 \\
    - \Brac[s]{s_k + s_l - 1 + \frac{\alphak}{2C} - \frac{1}{\vartheta} + \frac{\betal}{2\delta\vartheta^2}} & \Delta
    - c(\bm{\alpha}, \bm{\beta}).
  \end{align*}

  The optimal solution~$\Delta^{\star}$ is given by~\eqref{eq: Delta optimal}. We know that the optimal~$\delta^*$ is given by~\eqref{eq: Pat dual quadratic optimal delta}, then
  \begin{equation*}
    \delta^*
      = \sqrt{\frac{1}{4\vartheta^2 \ntil \tau} \Brac{\sum_{j\neq \hat{l}} \beta_j^2 + (\betal + \Delta^\star)^2}}
      = \sqrt{\delta^2 + \frac{1}{4\vartheta^2 \ntil \tau} (\Delta^{\star2} + 2\Delta^\star \betal)}.
  \end{equation*}
\end{proof}

\begin{proposition}[Update rule~\eqref{eq: update rule b,b} for problem~\eqref{eq: Pat dual quadratic}]\label{thm: patmat family quadratic update b,b}
  Consider problem~\eqref{eq: Pat dual quadratic}, update rule~\eqref{eq: update rule b,b} indices~$\npos + 1 \leq k \leq \ntil$ and~$\npos + 1 \leq l \leq \ntil$ and Notation~\ref{not: dual update rules}. Then the optimal solution~$\Delta^{\star}$ is given by~\eqref{eq: Delta optimal} where
  \begin{align*}
    \Delta_{lb} & = - \betak, \\
    \Delta_{ub} & = \betal, \\
    \gamma      & = -\frac{s_k - s_l + \frac{1}{2\delta \vartheta^2}(\betak - \betal)}{\K_{kk} + \K_{ll} - \K_{kl} - \K_{lk} + \frac{1}{\delta \vartheta^2}}, \\
    \delta^{\star}  & = \sqrt{\delta^2 + \frac{1}{2 \vartheta^2 \ntil \tau}({\Delta^{\star}}^2 + \Delta^{\star} (\betak - \betal))}.
  \end{align*}
\end{proposition}

\begin{proof}
  Constraint~\eqref{eq: Pat dual quadratic c1} is always satisfied from the definition of the update rule~\eqref{eq: update rule b,b}. Constraint~\eqref{eq: Pat dual quadratic c2} is also always satisfied since no~$\alpha_i$ is updated. Constraint~\eqref{eq: Pat dual quadratic c3} reads
  \begin{align*}
    0 \leq \betak + \Delta
    & \quad \implies \quad
    - \betak \leq \Delta, \\
    0 \leq \betal - \Delta
    & \quad \implies \quad
    \Delta \leq \betal, \\
  \end{align*}
  which gives the lower and upper bound of~$\Delta.$
  
  Using the update rule~\eqref{eq: update rule b,b}, objective function~(\ref{eq: Pat dual quadratic L1}-\ref{eq: Pat dual quadratic L2}) can be rewritten as a quadratic function with respect to~$\Delta$ as
  \begin{equation*}
    - \frac{1}{2} \Brac[s]{\K_{kk} + \K_{ll} - \K_{kl} - \K_{lk} + \frac{1}{\delta\vartheta^2}} \Delta^2
    - \Brac[s]{s_k - s_l + \frac{1}{2\delta\vartheta^2}(\betak - \betal)} \Delta
    - c(\bm{\alpha}, \bm{\beta}).
  \end{equation*}
  
  The optimal solution~$\Delta^{\star}$ is given by~\eqref{eq: Delta optimal}.   We know that the optimal~$\delta^*$ is given by~\eqref{eq: Pat dual quadratic optimal delta}, then
  \begin{equation*}
    \delta^*
      = \sqrt{\frac{1}{4\vartheta^2 \ntil \tau} \Brac{\sum_{j \notin \{\hat{l}, \hat{k}\}} \beta_j^2 + (\betak + \Delta^\star)^2 + (\betal - \Delta^\star)^2}} 
      = \sqrt{\delta^2 + \frac{1}{2\vartheta^2 \ntil \tau} (\Delta^{\star2} + \Delta^\star (\betak - \betal))}.
  \end{equation*}
\end{proof}

\subsubsection{Initialization}

As in the case of problem~\eqref{eq: toppushk family dual}, all update rules~\eqref{eq: update rules} assume that the current solution~$\bm{\alpha},$~$\bm{\beta},$~$\delta$ is feasible. So to create an iterative algorithm that solves problem~\eqref{eq: Pat dual hinge} or~\eqref{eq: Pat dual quadratic}, we need to have a way how to obtain an initial feasible solution. Such a task can be formally written as a projection of random variables~$\bm{\alpha}^0,$~$\bm{\beta}^0,$~$\delta^0$ to the feasible set of solutions
\begin{mini}{\bm{\alpha}, \bm{\beta}, \delta}{
  \frac{1}{2} \norm{\bm{\alpha} - \bm{\alpha}^0}^2
  + \frac{1}{2} \norm{\bm{\beta} - \bm{\beta}^0}^2
  + \frac{1}{2} (\delta - \delta^0)^2
  }{\label{eq: patmat family initialization}}{}
  \addConstraint{\sum_{i = 1}^{\npos} \alpha_i}{= \sum_{j = 1}^{\ntil} \beta_j}
  \addConstraint{0 \leq \alpha_i}{\leq C_1, \quad i = 1, 2, \ldots, \npos}
  \addConstraint{0 \leq \beta_j}{\leq C_2 \delta, \quad j = 1, 2, \ldots, \ntil,}
  \addConstraint{\delta }{\geq 0,}
\end{mini}
where the upper bounds in the second and third constraints depend on the used surrogate function and are defined as follows
\begin{align*}
  C_1 & = \begin{cases}
    C & \text{for hinge loss}, \\
    +\infty & \text{for quadratic hinge loss},
  \end{cases} &
  C_2 & = \begin{cases}
    \vartheta & \text{for hinge loss}, \\
    +\infty & \text{for quadratic hinge loss}.
  \end{cases}
\end{align*}

We show the way how to solve~\eqref{eq: patmat family initialization} only for hinge loss, since it is  trivial to solve it for quadratic hinge. Again, we will follow the same approach as in~\cite{adam2020projections} to solve this optimization problem. In the following theorem, we show that problem~\eqref{eq: patmat family initialization} can be written as a system of two equations of two variables~$\lambda$ and~$\mu.$ The theorem also shows the concrete form of feasible solution~$\bm{\alpha},$~$\bm{\beta},$~$\delta$ that depends only on~$\lambda$ and~$\mu.$

\begin{theorem}\label{thm: patmat family initialization}
  Consider problem~\eqref{eq: patmat family initialization} and some initial solution~$\bm{\alpha}^0,$~$\bm{\beta}^0$ and~$\delta^0.$ Then if the following condition holds
  \begin{equation}\label{eq: patmat family init alg condition}
    \delta^0 \le - C_2 \sum_{j = 1}^{\ntil} \clip[u]{0}{+\infty}{\beta_j^0 + \max_{i=1,\dots,\npos} \alpha_i^0}.
  \end{equation}
  the optimal solution of~\eqref{eq: patmat family initialization} amounts to~$\bm{\alpha} = \bm{\beta} = \bm{0}$ and~$\delta^0 = 0.$ In the opposite case, the following system of two equations
  \begin{subequations}\label{eq: patmat family init alg}
    \begin{align}
    0
      & = \sum_{i=1}^{\npos} \clip{0}{C_1}{\alpha_i^0 - \lambda} - \sum_{j=1}^{\ntil} \clip{0}{\lambda + \mu}{\beta_j^0 + \lambda},
    \label{eq: patmat family init alg 1} \\
    \lambda
      & = C_2 \delta^0 + C_2^2 \sum_{j=1}^{\ntil} \clip[u]{0}{+\infty}{\beta_j^0 - \mu} - \mu.
    \label{eq: patmat family init alg 2}
    \end{align}
  \end{subequations}
  has a solution $(\lambda,\mu)$ with $\lambda+\mu>0$ and the optimal solution of~\eqref{eq: patmat family initialization} is equal to
  \begin{align*}
    \alpha_i & = \clip{0}{C_1}{\alpha_i^0 - \lambda}, \\
    \beta_j & = \clip{0}{\lambda + \mu}{\beta_j^0 + \lambda}, \\
    C_2 \delta &= \lambda + \mu.
  \end{align*}
\end{theorem}

\begin{proof}
  The Lagrangian of~\eqref{eq: patmat family initialization} reads
  \begin{align*}
    \mathcal{L}(\bm{\alpha}, \bm{\beta}; \lambda, \bm{p}, \bm{q}, \bm{u}, \bm{v})
      = \frac{1}{2} \norm{\bm{\alpha} - \bm{\alpha}^0}^2
      + \frac{1}{2} \norm{\bm{\beta} - \bm{\beta}^0}^2
      + \frac{1}{2} (\delta - \delta^0)^2
     + \lambda \Brac{\sum_{i = 1}^{\npos} \alpha_i - \sum_{j = 1}^{\ntil} \beta_j} \\
     - \sum_{i = 1}^{\npos} p_i \alpha_i
     + \sum_{i = 1}^{\npos} q_i (\alpha_i - C_1)
     - \sum_{j = 1}^{\ntil} u_j \beta_j
     + \sum_{j = 1}^{\ntil} v_j (\beta_j - C_2 \delta).
  \end{align*}
  The KKT conditions then amount to the optimality conditions
  \begin{subequations}\label{eq:problem2_KKT}
    \begin{align}
      \frac{\partial \mathcal{L}}{\partial \alpha_i}
        & = \alpha_i - \alpha_i^0 + \lambda - p_i + q_i = 0,
        && \quad i = 1, 2, \ldots, \npos, \label{eq:problem2_KKT_opt1} \\
      \frac{\partial \mathcal{L}(\cdot)}{\partial \beta_j}
        & = \beta_j - \beta_j^0 - \lambda - u_j + v_j = 0,
        && \quad j = 1, 2, \ldots, \ntil, \label{eq:problem2_KKT_opt2} \\
      \frac{\partial \mathcal{L}(\cdot)}{\partial \delta}
        & = \delta - \delta^0 - C_2 \sum_{j = 1}^{\ntil} v_j = 0,
        \label{eq:problem2_KKT_opt3}
    \end{align}
  the primal feasibility conditions~\eqref{eq: patmat family initialization}, the dual feasibility conditions~$\lambda \in \R$, $p_i \ge 0$, $q_i\ge0$, $u_j \ge 0$, $v_j \ge 0$ and finally the complementarity conditions
  \begin{align}
    p_i \alpha_i & = 0,
      && \quad i = 1, 2, \ldots, \npos, \label{eq:problem2_KKT_comp1} \\
    q_i \Brac{\alpha_i - C_1} & = 0,
      && \quad i = 1, 2, \ldots, \npos, \label{eq:problem2_KKT_comp2} \\
    u_j \beta_j & = 0,
      && \quad j = 1, 2, \ldots, \ntil, \label{eq:problem2_KKT_comp3} \\
    v_j \Brac{\beta_j - C_2 \delta} & =0,
      && \quad j = 1, 2, \ldots, \ntil. \label{eq:problem2_KKT_comp4}
  \end{align}
  \end{subequations}

  \paragraph*{Case 1:} The first case concerns when the optimal solution satisfies~$\delta=0$. From the primal feasibility conditions, we immediately get~$\alpha_i = 0$ for all~$i$ and~$\beta_j = 0$ for all~$j$. Then~\eqref{eq:problem2_KKT_comp2} implies~$q_i=0$ and all complementarity conditions are satisfied. Moreover,~\eqref{eq:problem2_KKT_opt1} implies for all~$i$
  \begin{equation*}
    \lambda = \alpha_i^0 + p_i.
  \end{equation*}
  Since the only condition on~$p_i$ is the non-negativity, this implies~$\lambda \ge \max_i \alpha_i^0$.
  
  Similarly, from~\eqref{eq:problem2_KKT_opt2} we deduce
  \begin{equation*}
    v_j
      = \beta_j^0 +\lambda + u_j
      \ge \beta_j^0 + \lambda
      \ge \beta_j^0 + \max_{i=1,\dots,\npos} \alpha_i^0.
  \end{equation*}
  Since we also have the non-negativity constraint on $v_j$, this implies
  \begin{equation*}
    v_j \ge \clip[u]{0}{+\infty}{\beta_j^0 + \max_{i=1,\dots,\npos} \alpha_i^0}.
  \end{equation*}
  
  Condition~\eqref{eq:problem2_KKT_opt3} implies
  \begin{equation*}
    \delta^0
      = -C_2 \sum_{j = 1}^{\ntil} v_j
      \le -C_2 \sum_{j = 1}^{\ntil} \clip[u]{0}{+\infty}{\beta_j^0 + \max_{i=1,\dots,\npos} \alpha_i^0},
  \end{equation*}
  which is precisely condition~\eqref{eq: patmat family init alg condition}.

  \paragraph*{Case 2:} If~\eqref{eq: patmat family init alg condition} holds true, then from the discussion above we obtain that the optimal solution satisfies~$\delta > 0$. For any fixed~$i$, the standard trick is to combine the optimality condition~\eqref{eq:problem2_KKT_opt1} with the primal feasibility condition~$0 \le \alpha_i \le C_1$, the dual feasibility conditions~$p_i \ge 0$, $q_i \ge 0$ and the complementarity conditions~(\ref{eq:problem2_KKT_comp1}, \ref{eq:problem2_KKT_comp2}) to obtain
  \begin{equation}\label{eq:problem2_alpha}
    \alpha_i = \clip{0}{C_1}{\alpha_i^0 - \lambda}.
  \end{equation}

  Similarly for any fixed~$j$, we combine the optimality condition~\eqref{eq:problem2_KKT_opt2} with the primal feasibility condition~$0 \le \beta_j \le C_2 \delta$, the dual feasibility conditions~$u_j \ge 0,$ $v_j \ge 0$ and the complementarity conditions~(\ref{eq:problem2_KKT_comp3}, \ref{eq:problem2_KKT_comp4}) to obtain
  \begin{align}
    \beta_j & = \clip{0}{C_2 \delta}{\beta_j^0 + \lambda}, \label{eq:problem2_beta} \\
    v_j & = \clip[u]{0}{+\infty}{\beta_j^0 + \lambda - C_2 \delta}. \label{eq:problem2_rho}
  \end{align}

  Note that we now obtain the following system
  \begin{align*}
    \sum_{i=1}^{\npos} \clip{0}{C_1}{\alpha_i^0 - \lambda} - \sum_{j = 1}^{\ntil} \clip{0}{C_2 \delta}{\beta_j^0 + \lambda}
      & = 0, \\
    \delta - \delta^0 - C_2 \sum_{j = 1}^{\ntil} \clip[u]{0}{+\infty}{\beta_j^0 + \lambda - C_2 \delta}
      & = 0.
  \end{align*}
  Here, the first equation follows from plugging~\eqref{eq:problem2_alpha} and~\eqref{eq:problem2_beta} into the feasibility condition~$\sum_i \alpha_i =\sum_j \beta_j$ while the second equation follows from plugging~\eqref{eq:problem2_rho} into~\eqref{eq:problem2_KKT_opt3}. Finally, system~\eqref{eq: patmat family init alg} follows after making the substitution $C_2 \delta = \lambda + \mu$.
\end{proof}

System~\eqref{eq: patmat family init alg} is relatively simple to solve, since equation~\eqref{eq: patmat family init alg 2} provides an explicit formula for~$\lambda$. Let us denote it as $\lambda(\mu)$, then we denote the right-hand side of~\eqref{eq: patmat family init alg 1} as
\begin{equation}\label{eq: patmat family init alg h}
  h(\mu) :=
    \sum_{i=1}^{\npos} \clip{0}{C_1}{\alpha_i^0 - \lambda(\mu)} - \sum_{j=1}^{\ntil} \clip{0}{\lambda(\mu) + \mu}{\beta_j^0 + \lambda(\mu)}.
\end{equation}
Then the system of equations~\eqref{eq: patmat family init alg} is equivalent to solving~$h(\mu) = 0$. The following lemma states that~$h$ is a non-decreasing function in~$\mu$ on~$(0,\infty)$ and thus the equation~$h(\mu) = 0$ is simple to solve using any root-finding method. Note that if~$\delta^0 < 0$, then it may happen that~$\lambda + \mu < 0$ if the initial~$\mu$ is chosen large. In such a case, it suffices to decrease~$\mu$ until~$\lambda + \mu$ is positive.

\begin{lemma}\label{lemma: patmat family initialization h}
  Function $h$ is non-decreasing in~$\mu$ on~$(0,\infty)$.
\end{lemma}

\begin{proof}[Proof of Lemma~\ref{lemma: patmat family initialization h} on page~\pageref{lemma: patmat family initialization h}]
  Consider any~$\mu_1 < \mu_2$. Then from~\eqref{eq: patmat family init alg 2} we obtain both~$\lambda(\mu_1) \ge \lambda(\mu_2)$ and~$\mu_1+\lambda(\mu_1) \ge \mu_2 + \lambda(\mu_2)$. The statement then follows from the definition of~$h$ in~\eqref{eq: patmat family init alg h}.
\end{proof}

\end{document}